# Advances in LLMs with Focus on Reasoning, Adaptability, Efficiency, and Ethics

*Asifullah Khan[1,2,3*], Muhammad Zaeem Khan[2], Aleesha Zainab[2], Saleha Jamshed[2], Sadia Ahmad[2], Kaynat Khatib[2], Faria Bibi[2], and Abdul Rehman[1,4].*

[1]Pattern Recognition Lab, DCIS, PIEAS, Nilore, Islamabad, 45650, Pakistan.
[2]PIEAS Artificial Intelligence Center (PAIC), PIEAS, Nilore, Islamabad, 45650, Pakistan.
[3]Deep Learning Lab, Center for Mathematical Sciences, PIEAS, Nilore, Islamabad, 45650, Pakistan.
[4]Department of Biomedical Engineering, The Chinese University of Hong Kong, Shatin, NT, Hong Kong.

***Corresponding author(s) E-mail(s):** asif@pieas.edu.pk;

## Abstract

This survey paper outlines the key developments in the field of Large Language Models (LLMs), including enhancements to their reasoning skills, adaptability to various tasks, increased computational efficiency, and the ability to make ethical decisions. The techniques that have been most effective in bridging the gap between human and machine communications include the Chain-of-Thought prompting, Instruction Tuning, and Reinforcement Learning from Human Feedback. The improvements in multimodal learning and few-shot or zero-shot techniques have further empowered LLMs to handle complex jobs with minor input. A significant focus is placed on efficiency, detailing scaling strategies, optimization techniques, and the influential Mixture-of-Experts (MoE) architecture, which strategically routes inputs to specialized subnetworks to boost predictive accuracy, while optimizing resource allocation.

This survey also offers a broader perspective on recent advancements in LLMs, going beyond isolated aspects such as model architecture or ethical concerns. Additionally, it explores the role of LLMs in Agentic AI and their use as Autonomous Decision-Making Systems, and categorizes emerging methods that enhance LLM reasoning, efficiency, and ethical alignment. The survey also identifies underexplored areas such as interpretability, cross-modal integration, and sustainability. While significant advancements have been made in LLMs, challenges such as high computational costs, biases, and ethical risks remain. Overcoming these requires a focus on bias mitigation, transparent decision-making, and explicit ethical guidelines. Future research will generally focus on enhancing the model's ability to handle multiple inputs, thereby making it more intelligent, safe, and reliable.

## 1. Introduction

Over the last 2 decades, artificial intelligence has incorporated the notion of Large Language Models (LLMs). This technology is unique and is able to facilitate the reading and writing of hundreds of human languages by machines. From the development of smart assistants to the generation of web content and the translation of documents in different languages, LLMs can perform any task in modern AI. For the development of these deep learning models, a neural network structure called a transformer is used. They are very dependent on a lot of data and computational resources in order to understand human language, generate text and answer comprehension questions.

These models have recently become very successful because of their ability to perform several complex tasks related to language like the generation of text that is reasonable in context and flawless. Vaswani et al. (2017) created a model called “Attention is All You Need,” showcasing a novel transformer architecture which revolutionized the field of LLMs and created a paradigm shift within deep learning and natural language processing [1].

In the recent past, the advancements within LLMs have been phenomenal, with the rise of new generations, LLMs have begun to achieve new heights within the realms of scale, reasoning and multimodal integration. The advances within LLMs from GPT-3 (2020) to GPT-5 (2025) and its contemporaries Claude (Anthropic), Gemini (Google DeepMind), DeepSeek (China) and LLaMA (Meta) have been mapped out in the chronological Figure 1 and showcases the distinct research directions shaping Generative AI to become more efficient and safer with an improved ethical alignment.

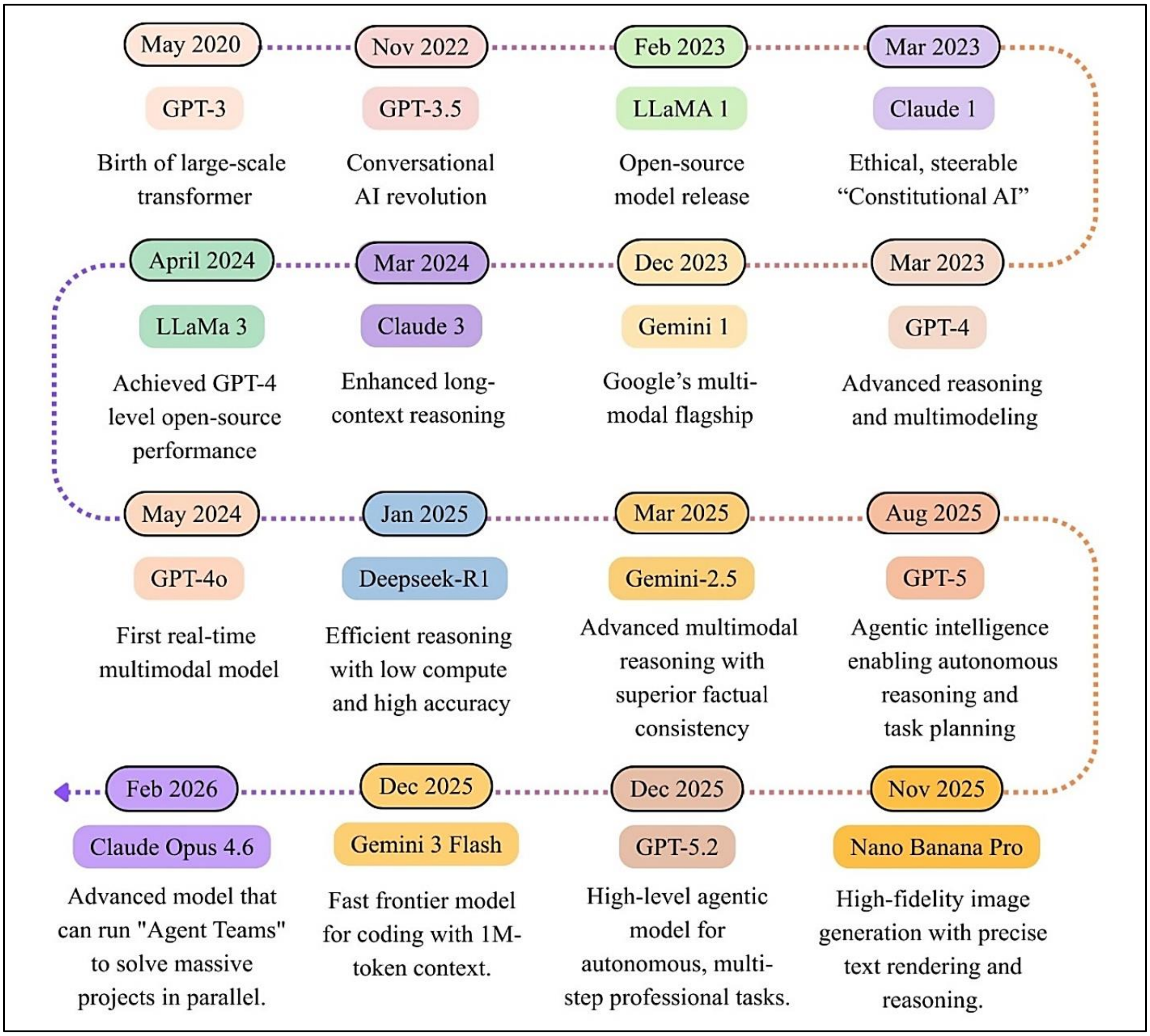

**Chronological timeline illustrating the evolution of large language models (LLMs) from GPT-3 (2020) to frontier agentic systems in early 2026. The progression includes major contemporaries such as Claude, Gemini, DeepSeek, and LLaMA. The later stages reflect the emergence of agent-native architectures, persistent memory, multi-agent collaboration, and autonomous task planning, marking the transition from conversational AI to unified, tool-using foundation agents [73].**

As with most technologies, LLMs have their cons. There often lies a trade-off in computational capacity vs. accuracy. The development and deployment of LLMs is continuously hampered by hallucinations, biases within and ethical implications of the data, lack of reasoning and the true capabilities of the model. These issues need to be resolved to ensure LLMs are used responsibly.

Researchers have discovered approaches like Chain-of-Thought (CoT) prompting and Instruction Tuning to handle these limitations. These methods allow us to engage in multi-step reasoning before responding, which can improve how well models follow human-like instructions. Additionally, with multimodal models, LLMs can learn from various inputs like audio and visuals, thus opening opportunities beyond just text.

Fairness and transparency in LLMs are major research areas. Ethical AI and bias mitigation aim to reduce biases in the training dataset so that AI systems can be morally responsible. Also, learning paradigms, like Few-Shot and Zero-Shot Learning, permit the models to generalize from few instances and thus perform better in real life. Methods such as Reinforcement Learning from Human Feedback (RLHF) could be used to fine-tune the answers given by models from the feedback by the user, thus making them more in line with human expectations. As LLMs continue to scale, efforts to improve efficiency become increasingly important. While at it, Self-Supervised Learning uses massive portions of unlabeled data to increase model generalization.

Whereas the existing surveys tend to concentrate on the aspects of Large Language Models (LLMs), such as architecture or ethical issues, this survey provides a broader perspective of recent developments. It classifies emerging methods that improve LLM reasoning, effectiveness, and ethical inclination. In doing so, the survey not only guides newcomers through the complex landscape of the field but also helps experienced researchers identify trends, connections, and open issues within the subfields.

In addition, this survey evidences the lack of attention to interpretability in the models, the cross-modal integration, and sustainability; this aspect of the latest research practice has been under-explored in the literature. It also touches on the interaction of the LLMs with such technologies as reinforcement learning and multimodal models, indicating trends for further research for more robustness, fairness, and societally aligned functioning. The survey does not cover older, well-established methods discussed at great length in the previous surveys. It focuses on the most recent and influential work that will determine future development work in LLM.

This paper is structured as follows: Sections 2 and 3 study critical thinking strategies that enhance LLM's competency in conducting logical thinking and adhering to human instructions: Chain-of-Thought (CoT), Prompting, and Instruction Tuning. Section 4 covers ethical considerations concerning Ethical AI and Bias Mitigation to ensure fairness and transparency in AI-driven applications. Section 5 discusses Multimodal Models, which can combine and transform various data types other than text. Section 6 leads to Few-Shot and Zero-Shot Learning, which educates us in how LLMs generalize to new tasks, with few training data, and Section 7 describes Reinforcement Learning from Human Feedback (RLHF). Section 8 refers to Model Scaling and Efficiency, the task of determining the ideal size and performance of the model based on its application's needs. Section 9 explains how LLM-based agents mimic human-like autonomy in dynamic environments. Section 10 explains how we can shift from a monolithic model to select multiple expert models. Sections 11 and 12 discuss the challenges and future studies in LLM.

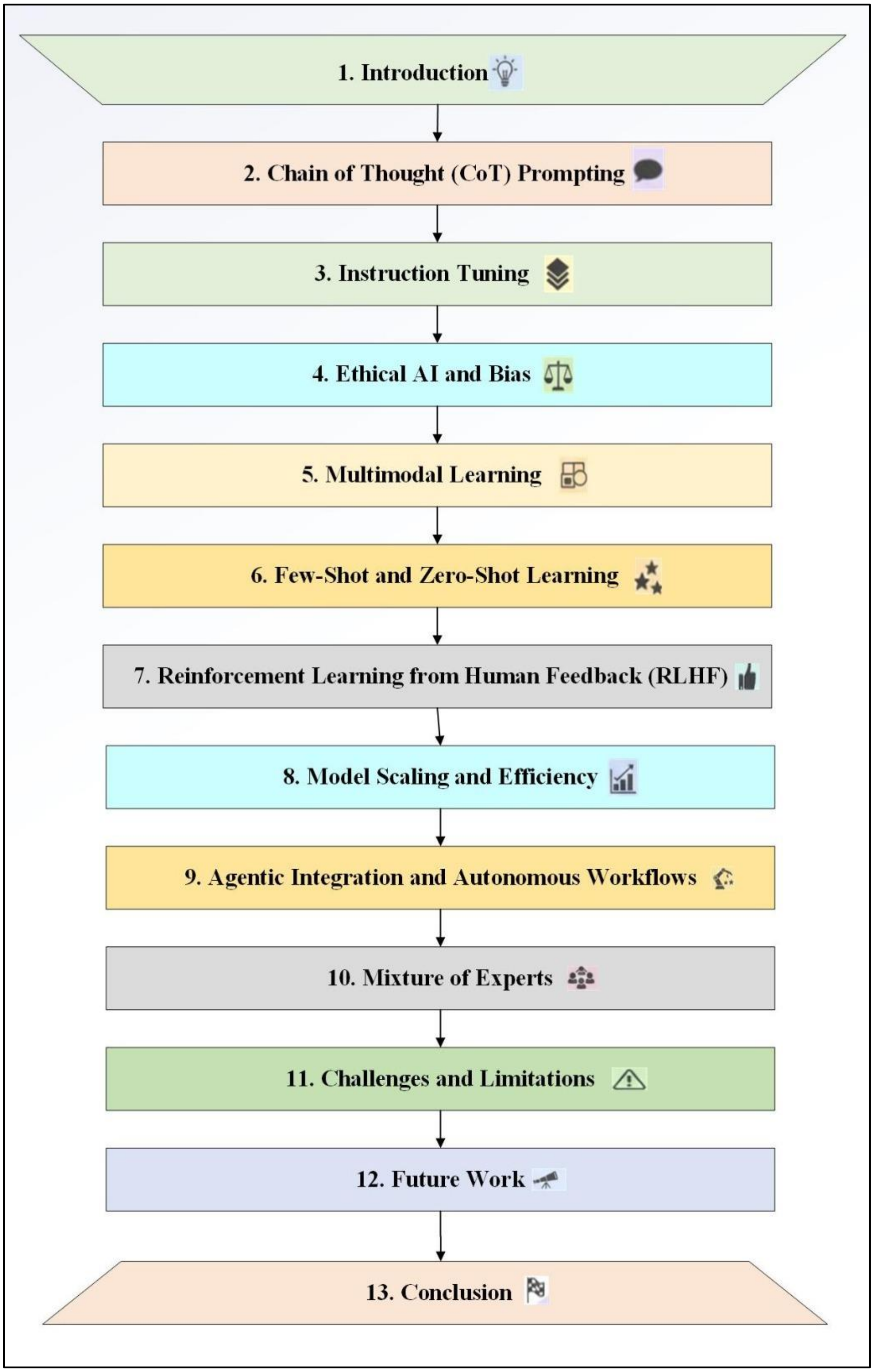

**Figure 2: Layout of different sections of the survey paper**

## Abbreviations

| Abbreviations | Full Form |
|---|---|
| **BERT** | Bidirectional Encoder Representations from Transformers |
| **BLIP** | Bootstrapped Language-Image Pretraining |
| **CLIP** | Contrastive Language-Image Pretraining |
| **CoT** | Chain-of-Thought |
| **FLAN** | Fine-tuned Language Net |
| **GDPR** | General Data Protection Regulation |
| **GPU** | Graphics Processing Unit |
| **GPT** | Generative Pre-trained Transformer |
| **FL** | Federated Learning |
| **IoT** | Internet of Things |
| **LLM** | Large Language Model |
| **LLaMA** | Large Language Model Meta AI |
| **ML** | Machine Learning |
| **MLLM** | Multimodal Large Language Model |
| **NLP** | Natural Language Processing |
| **PaLM** | Pathways Language Model 2 (developed by Google) |
| **QML** | Quantum Machine Learning |
| **RLHF** | Reinforcement Learning from Human Feedback |
| **SSL** | Self-Supervised Learning |
| **T5** | Text-to-Text Transfer Transformer |
| **TPU** | Tensor Processing Unit |
| **XAI** | Explainable AI |
| **ZSL** | Zero-Shot Learning |

## 2. Chain-of-Thought (CoT) Prompting

Chain-of-thought (CoT) prompting is an emerging technique used by recent large language models (LLMs) [2]. In this technique, the model performs intermediate reasoning steps before a final answer is provided, unlike standard input-to-output mapping. By dividing complicated problems into smaller, easy-to-solve logical steps, resolving them one by one, and reaching the result, CoT makes difficult activities easier to process, just like human thought.

CoT proves to be excellent in complex tasks, such as arithmetic, common sense, and logical deduction [3]. In addition, CoT introduces transparency between the user and the answer, as the users can see intermediate steps. This approach is now used to strengthen the reasoning abilities of LLMs in various domains, making them more viable for practical use.

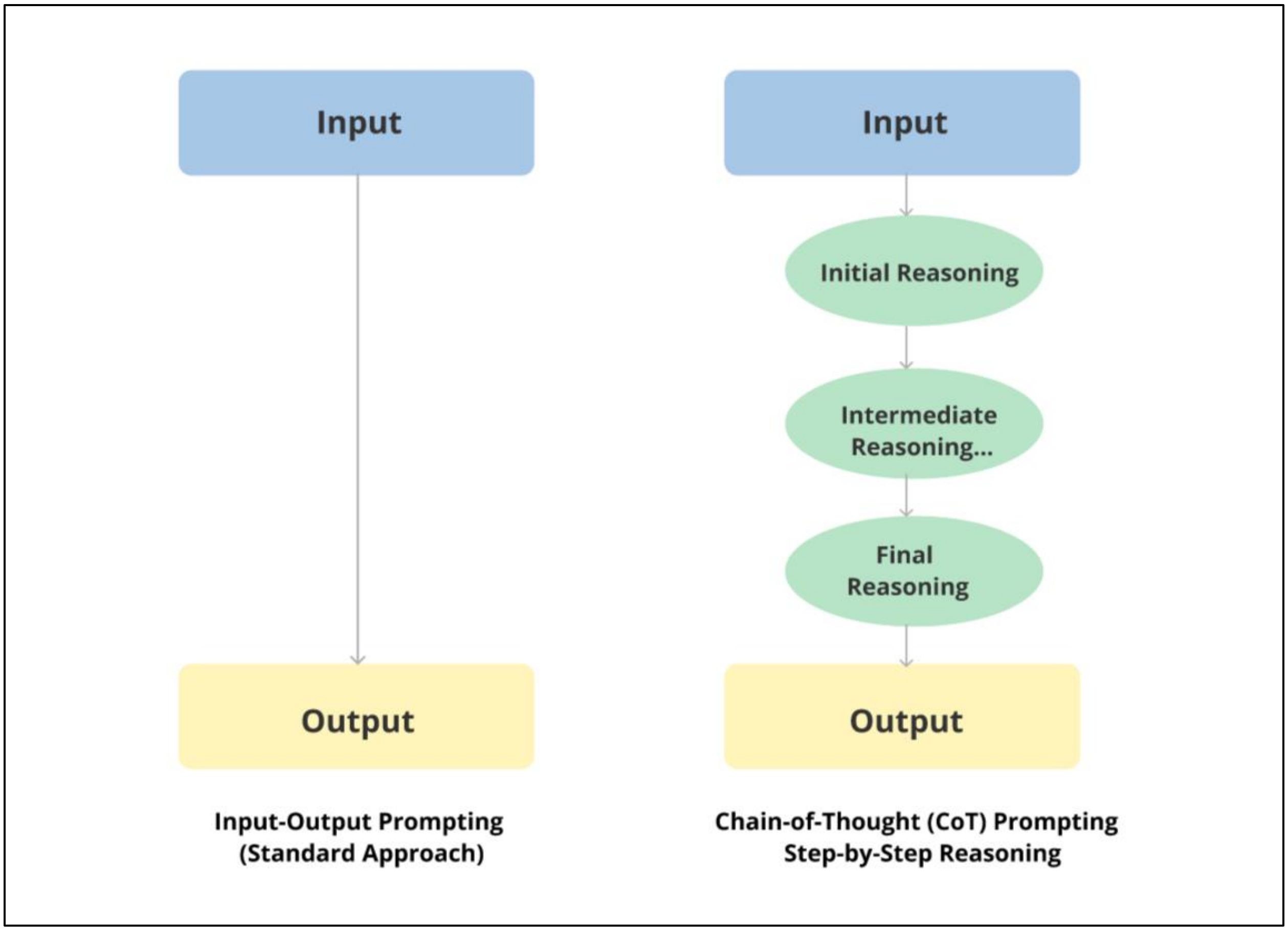

**Figure 3: Workflow of standard input-output prompting and Chain-of-Thought (CoT) prompting.**

## 2.1 Types of Chain of Chain-of-Thought

While traditional CoT relies on explicit examples within the prompt, several refined approaches have been developed to enhance model performance:

**Zero-Shot CoT:** By beginning the prompt with a phrase such as "Let's think step by step," the model is prompted to generate its internal reasoning sequence without any prior examples. This technique eliminates the need for manually supplied demonstrations.

**Few-Shot CoT:** The user includes one or more complete examples that detail the question, the intermediate reasoning (the "thought chain"), and the final answer. These examples guide the model toward the desired reasoning pattern.

**Self-Consistency:** An advanced method in which the model produces multiple distinct reasoning paths for the same problem. The final answer is selected via a majority vote across all generated paths, leading to a notable improvement in mathematical accuracy.

## 2.2 Examples/Use Cases

CoT resolves mathematical questions logically. Instead of providing a direct answer, it solves arithmetic problems step by step. CoT effectively solves the reasoning problems because common sense is a matter of logic, and the mixing of facts or inferring.

CoT prompting is suitable for the summary and analysis of text. The model can do this because it converts a long text into several smaller parts and summarizes each part individually before combining them, producing a consistent summary. This method of summarizing material ensures comprehensive and systematic results, making it most appropriate for processing long documents, reports, or articles.

CoT is also valuable for code generation. It divides the task into several sub-tasks, resolves each of them, and synthesizes them into a final code, which increases readability and helps the users understand the logic behind the solution.

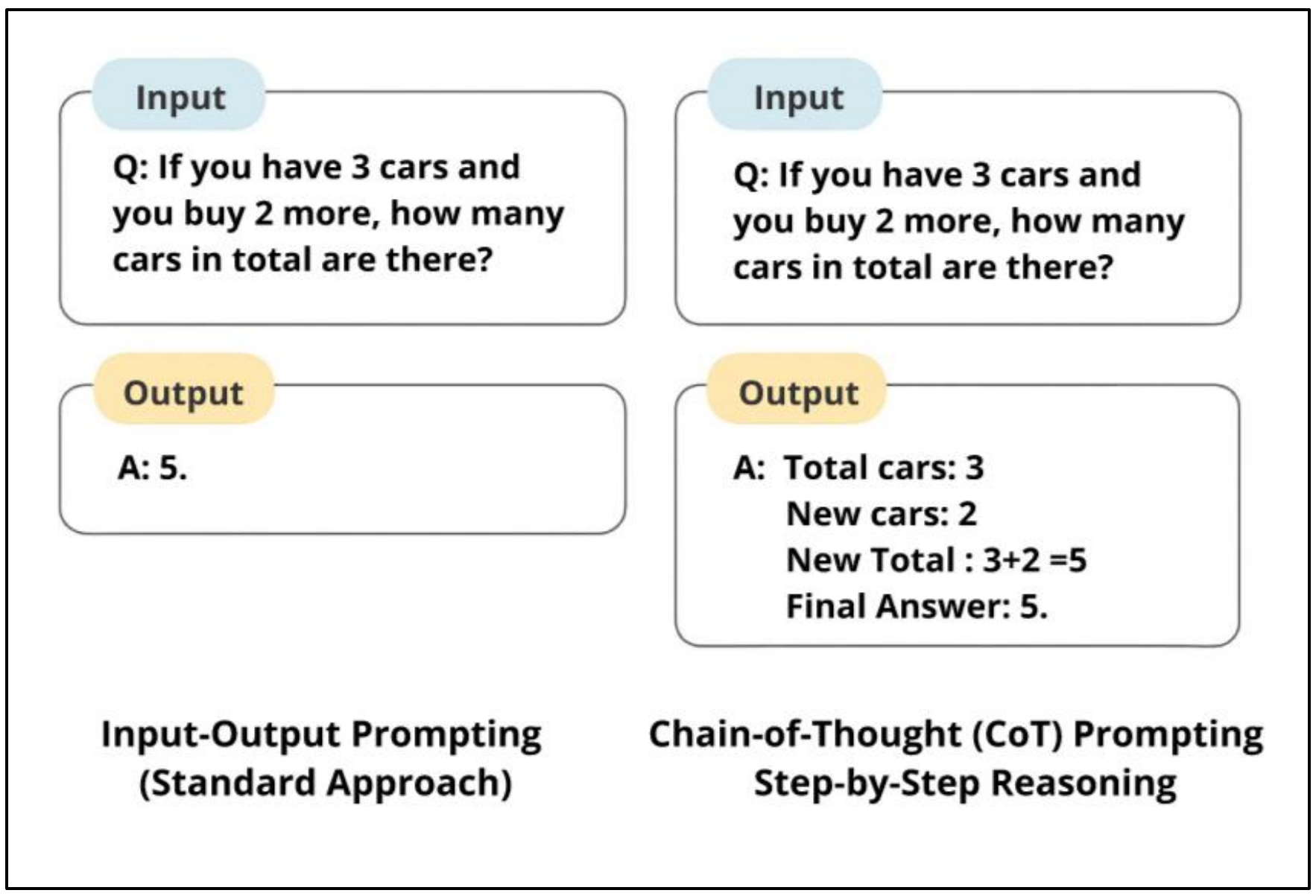

**Figure 4: Example of input-output prompting versus Chain-of-Thought (CoT) prompting.**

## 2.3 Practical Implications

AI systems based on CoT can assist students by using them as virtual tutors. They could break down complicated concepts into simple steps that are easier to understand, through which they could provide comprehension. With the help of CoT, the AI models can resolve multi-step math word problems by dividing every calculation step, making the reasoning process more straightforward and effective. Lawyers can use CoT to analyze legal documents and gain insights into complex legal interpretations. It can support them in terms of finding the potential problems and contradictions in the policies and laws. Logical thinking at every step can assist them in developing stronger arguments. AI can help businesses make better decisions and work in many areas, including financial analysis, customer relationships, and risk management.

As a result of a step-by-step analysis of symptoms, CoT may help healthcare professionals to produce an accurate diagnosis, hence preventing errors in diagnosis. It can also assist in giving personalized treatment recommendations and forecasting the disease progression. The developers can use CoT to locate mistakes in code. Developers can use CoT to identify errors in code because it provides a rationale for why each step should be carried out. By analyzing these steps, developers can discover what causes potential bugs. Combined with an AI model that automatically generates students' grades, CoT can have efficient results and explain the reasons for grades. It enhances transparency in the system of education [4].

## 2.4 CoT vs Direct Answer

| Aspect | CoT | Direct Answer |
|---|---|---|
| **Reasoning Approach** | Generates step-by-step reasoning before providing a final answer. | Provides an immediate response without showing its reasoning process. |
| **Accuracy and Reliability** | Enhances accuracy in tasks that involve multi-step solutions by decomposing complex problems into more manageable, logical steps. It improves the model's ability to solve multi-step problems. | It is accurate and efficient for simple questions, but doesn't perform well on complex tasks. |
| **Transparency and Explainability** | Explains how a solution is derived, making the process easier to understand and transparent. This transparency is highly valuable in education, legal analysis, and healthcare applications. | Direct answering is not transparent, which makes it difficult to trust the results produced by the model. |
| **Speed and Efficiency** | Produces step-by-step answers, which increases the model's response time and makes it slower than direct answering. | Models using direct answering provide a faster response, making them more suitable for real-time applications where speed is crucial. |
| **Suitability for Different Tasks** | CoT Prompting is most suitable for complex reasoning tasks like math problem-solving, code debugging, legal analysis, and medical diagnosis support [5]. | Direct answering is suitable where a quick and factual answer is required, such as answering simple questions, providing definitions, or retrieving specific information. |

## 2.5 Extended Reasoning Frameworks

In recent years, CoT has been expanded into more structured reasoning frameworks, improving the diversity and reliability of problem-solving techniques.

### 2.5.1 Tree-of-Thought (ToT)

Tree-of-Thought builds on CoT by enabling the model to explore multiple reasoning paths at the same time, similar to a search tree. Rather than following a single linear chain, the model branches into alternative reasoning paths, evaluates partial solutions, and selects the most promising one. This approach facilitates strategic exploration and decision optimization, making it effective in planning and complex reasoning tasks.

### 2.5.2 Graph-of-Thought (GoT)

Graph-of-Thought concept of CoT by connecting reasoning nodes in a non-linear, graph-like structure. Each node represents a partial idea or reasoning step, while edges capture dependencies between thoughts. GoT enables the model to combine various reasoning paths, revisit previous steps, and merge partial conclusions, providing a richer and adaptable reasoning process suitable for multi-domain knowledge integration.

### 2.5.3 Self-Verification and Reflection Loops

In self-verification, the model analyzes and evaluates its own reasoning chain, identifying inconsistencies or errors before producing the final answer. Reflection loops introduce a feedback stage, where the model reanalyzes its output based on its previous reasoning track, resulting in improved reliability and fewer hallucinations. These mechanisms transform CoT from a static reasoning pipeline into an iterative self-correcting process.

### 2.5.4 Multi-Agent Debate Methods

Multi-agent debate frameworks use multiple LLMs (or model instances) to argue and evaluate each other's reasoning paths. Each agent provides a justification for its proposed solution, while others identify weaknesses or contradictions. This organized discussion promotes diversity of opinion and convergence toward more accurate, well-justified outcomes. Better factual accuracy and interpretability in decision-critical applications by integrating debate with CoT or ToT.

# 3. Instruction Tuning

Supervised fine-tuning, or instruction tuning, enhances outcomes for large language models (LLMs) [6]. This specific form of learning involves teaching LLMs to perform a variety of tasks using labeled datasets where instructions are expressed in natural language alongside their resultant outputs [7]. This improves a model's capacity to follow directions and rules, not only within limited-scope tasks but in broader contexts. It broadens the model's ability to adapt and properly respond to prompts made by humans, enabling the model to understand correctly, thus increasing generalization.

The primary aim of instruction tuning is to expand the scope of a model's adaptability to the extent that compliance even in the presence of unfamiliar tasks is achieved. Multitasking becomes possible, and the overall end-user satisfaction increases during interactions with the model's functionalities.

This has been critical for the InstructGPT, FLAN, and T0 models, which have been designed to follow instructions and display robust generalization and greater alignment to user needs and preferences. Giving current benchmarks in natural language allows model flexibility and makes them stronger as general-purpose assistants, thus changing the outlook for instruction following the revolution in NLP. Therefore, it becomes a focal stage in creating large language models that are not only proficient with human values and expectations but seamlessly integrate them during task performance.

## 3.1 CoT -Focused Instruction Tuning (CoT -FT)

CoT-Focused Instruction Tuning (CoT-FT) is an advanced methodology that integrates training and prompting strategies. In this approach, large language models are explicitly fine-tuned on datasets that contain not only final answers but also the intermediate reasoning steps commonly referred to as "rationales" or "chains of thought." By incorporating rationale-target pairs into the training regimen, the model learns to internalize the logical decomposition required to solve a problem. Consequently, the model's zero-shot reasoning performance is substantially improved, enabling it to generate transparent, accurate, and step-wise solutions for complex, multi-step tasks without the need for explicit user-triggered reasoning cues.

## 3.2 Examples/Use Cases

Instruction tuning has been used in the major LLMs, and it is proven to be effective in different tasks. Examples of instruction tuning usage are as follows:

OpenAI's InstructGPT is a model that understands and follows user instructions. It uses a four-step procedure: fine-tuning the model at a large scale of tasks, generating a reward model, supervised fine-tuning, and reinforcement learning. This makes it more reliable in generating results that match human expectations. It can do many things, like summarizing whole articles, generating code snippets, writing essays, etc.

Google's FLAN is a language model fine-tuned for tasks like translation, summarization, and question answering. It has been trained on thousands of written instruction tasks and thus generalizes well to new tasks. For example, the model can translate a sentence from one language to another, solve mathematical functions from the prompt, and explain scientific notions in a non-scientific way.

Apart from those models, research has demonstrated the power of instruction tuning. A key paper, "Fine-Tuned Language Models Are Zero-Shot Learners," testified how instruction-tuned models could accomplish unseen tasks. According to this research, models acquired through training on varied instructions can generalize and perform unseen tasks, all via a prompt [8].

## 3.3 Practical Implications

Tuning of instruction contains a wide range of practical applications in numerous industries, assisting users to perform numerous tasks [9]. Below are critical use cases:

Instruction-tuned LLMs offer intelligent chatbots that understand customer queries and context-aware solutions and handle them accordingly based on sentiment. Companies employ them to automate support services, cut costs, and increase customer satisfaction with round-the-clock services. Using instruction-tuned LLMs, healthcare doctors can diagnose illnesses by analyzing symptoms and medical records. Virtual health assistants, which are LLM-driven, may remind patients of their medicine and offer individual treatment, answer health-oriented questions, and monitor chronic illnesses.

Instruction tuning enhances AI in education through the provision of personalized learning experiences. Teachers can also use this technology to create lesson plans, quizzes, and summaries. In marketing and media, instruction-tuned models help writers and marketers since they can produce a blog post, social media caption, video script, and advertisement based on specific prompts. Such models can vary in tone and style depending on the platform and the target audience, saving time and increasing creativity.

Developers apply instruction-tuned models to automate coding jobs, debug mistakes, and make it possible to explain complex programming concepts. LLMs can create whole text snippets out of descriptions, support writing docs, and offer solutions to coding problems, saving time for programmers and reducing mistakes.

## 3.4 Pre-training vs Instruction Tuning

| Aspect | Pre-training | Instruction Tuning |
| --- | --- | --- |
| **Definition** | In the initial training phase, the model is given a wide range of text data to learn general language patterns. | The next phase is when the model is fine-tuned with specific instructions and data to enhance its performance on tasks. |
| **Goal** | The main aim is to broadly develop an understanding of language, context, and general world knowledge. | To enhance the ability of models to follow human instructions and complete specific tasks accurately. |
| **Key Challenge** | Lacks task-specific focus; the model may not be effective with direct instructions or specific tasks. | The diversity and clarity of the instruction data determine success; poor examples can limit generalization. |
| **Techniques Used** | Unsupervised learning on massive text datasets of various sources. | Supervised fine-tuning with labeled instruction-response pairs, at times supplemented with human feedback. |
| **Impact on Accuracy** | It provides a wide range of functionality but lacks precision unless fine-tuned to specific tasks. | Improves the performance of a task as well as reliability on instruction-based prompts. |
| **Trade-off Consideration** | Focus on general capabilities instead of task-specific performance; it provides a strong foundation but is not specifically optimized for tasks. | Requires a delicate balance to avoid overfitting to seen tasks while improving performance on new(unseen) ones. |

## 4. Ethical AI and Bias Mitigation

Ethical AI analyzes AI systems and aims for them to be used in a fair and responsible manner. For instance, AI is important in almost every aspect of our lives. For instance, AI is important in the workplace, in loan applications, and in the healthcare system. AI, like any other new technology, has several pros and cons. One of the negative aspects is the potential for bias. This bias can be deliberate or structural and can be the result of discrimination, whether by gender or race, from poor or obsolete datasets [10], [11]. Moreover, even in the presence of a sufficient dataset, AI systems may develop certain patterns that lead to undesired inequitable outcomes [12].

All these factors must be addressed to make AI trustworthy to all users, hence making the technology reliable. Ethical AI is primarily responsible for accountability, considering any bias in processes, and aiming for equitable treatment across every level while maintaining a level of simplicity. On the other hand, responsibility entails developers, organizations, and companies bearing the brunt of the ethical results of their AI systems [13]. Bearing this in mind, researchers are in continuous pursuit of systems to grant technologies that aim towards mitigating bias to enhance our technologies.

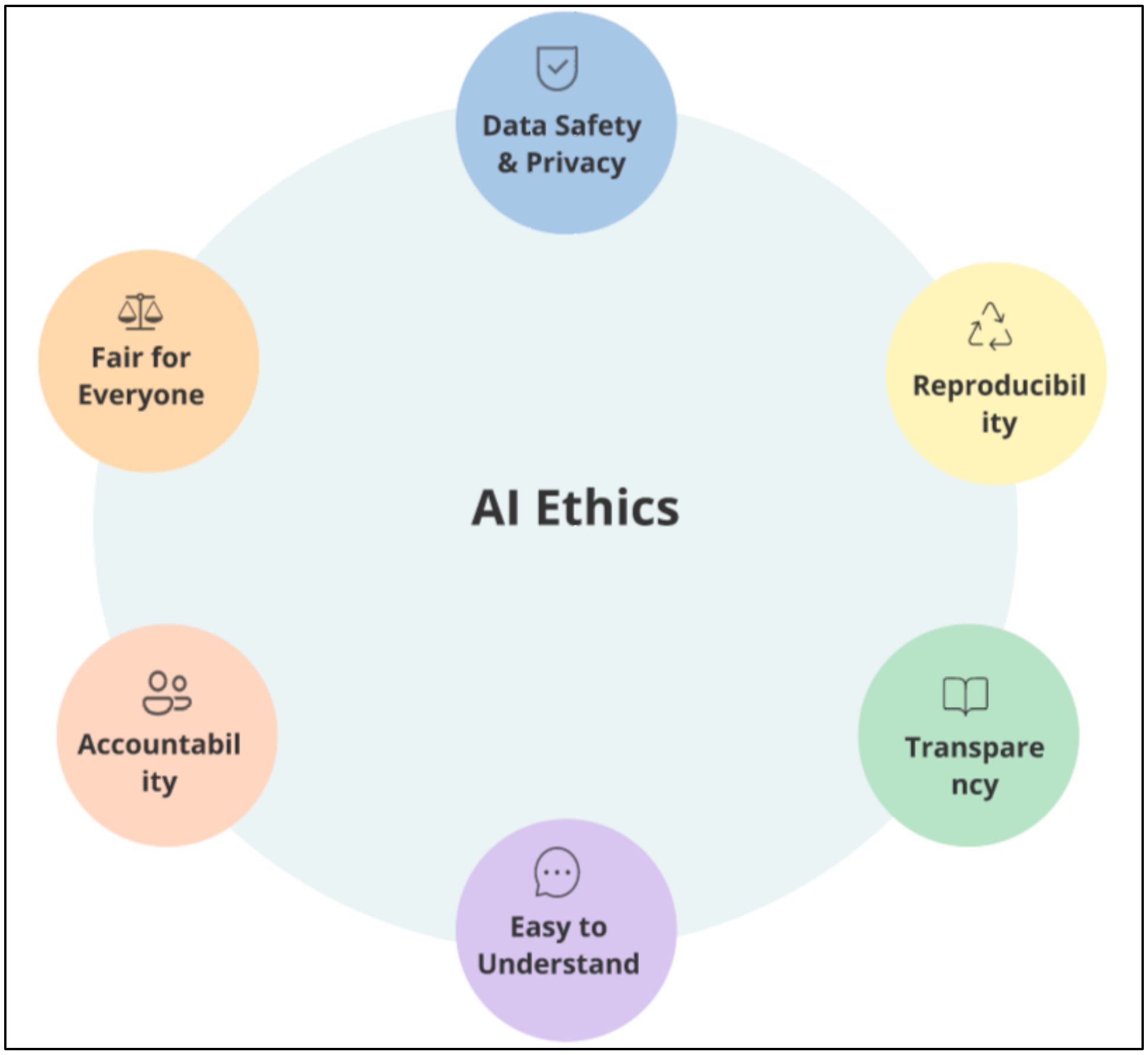

**Figure 5: Circular representation of AI ethics: privacy, fairness, accountability, transparency, explainability, reproducibility.**

## 4.1 Examples and Use Cases

Multiple tools and accompanying research seek to assist in bias detection and mitigation in AI and machine learning systems. Consider IBM's AI Fairness 360, an open-source application with tools dedicated to the detection and mitigation of bias within ML algorithms [14]. It assists developers and academics with balanced evaluations and offers solutions at different phases of AI training. Google's What-If Tool is equally valuable, providing graphing capabilities and model explainability to aid developers' understanding [15]. Developers can analyze model outputs across various demographic groups to determine how differential changes to input data can affect the overall results.

Similarly, Microsoft Fairlearn is also considered a bias mitigation tool but focuses on AI predicting cases and control groups within a relevant population [16]. It maintains some measures employed during the training of the models to ensure the AI is fair to all the groups being considered. Aequitas is another amazing tool developed by the University of Chicago [17]. The tool is publicly accessible as it supports fairness metrics for use in public healthcare prediction and criminal justice systems. The final section focuses on bringing AI decisions to a more comprehensible level through a tool called AI Explainability 360 Kit developed by IBM. It offers a variety of options to make AI decision-making processes more transparent so that the system is justifiable and responsible [18]. The more recent ones that include Knowledge Graph-Augmented Training (2025) add organized specific knowledge to combat biased results in high stakes cases.

Kumar et al. (2025) presents KGAT or Knowledge Graph-Augmented Training and he is the first to detect bias in the Large Language Model (LLM) and adopt a method to mitigate it. The technique reduces the bias output of the LLMs by adding systematic, field-based knowledge graphs during the training phase [19]. In the paper called "On the Dangers of Stochastic Parrots: Can Language Models Be Too Big?" by Bender et al. she focused on the more substantial datasets that these LLMs are trained on and pointed out that it is these datasets that bias the model and amplify the biases of it and she emphasized the need for the LLMs to have a bias mitigation technique and transparency [12]. The "Gender Shades" study by Buolamwini and Gebru also highlighted bias in facial recognition systems and pointed out that commercial AI models had much higher error rates for darker and female faces which led major companies to improve their algorithms.

## 4.2 Practical Implications

AI bias isn't just something that happens in computers; it affects real people in most areas, like jobs, law, healthcare, finance, and social media. That's why it's important to address these biases [10]. If AI is unfair, it may keep its unfair practices and not contribute to problem-solving. Companies apply AI to sort out job applications, but these systems sometimes discriminate against some groups. "For example, if past hiring data included mostly men in leadership roles, AI might continue to favor men over women [12]. By using fair AI techniques, we can ensure everyone has an equal chance of getting hired. Courts are starting to use AI to help decide bail and sentencing. But if the system learns from past legal cases where certain groups were subject to unfair treatment, it may continue that pattern. This bias can lead to unjust rulings, especially against minority communities [23]. Fixing bias in legal AI can make the system fairer for everyone. AI helps doctors diagnose diseases and suggest treatments, but it can be biased if trained on data that mostly comes from certain groups. For example, studies show some AI models miss diseases in women and people of color because most medical research focuses on white male patients [11]. Fair AI can ensure accurate healthcare for all. Banks use AI for loan and credit card decisions, but biased past data may lead to unfair rejection of certain ethnic groups or low-income individuals. AI can help make sure everyone gets a fair chance. In 2024/25, the *EU AI Act* formally classified credit scoring and loan approval as 'high-risk' systems, requiring fairness audits and explainability reports. Similarly, in 2025 the *U.S. SEC* proposed new rules obligating financial firms to document AI decision pipelines for compliance. AI helps social media platforms remove harmful content [24], but sometimes, it censors reasonable voices. Ethical AI can help keep things fair, to make online spaces safer and more inclusive of everyone's voice. AI is already used in schools to grade essays and assignments, but if the system is biased, it may choose some students over others, even though the submitted work is good [25].

### 4.3 Bias Mitigation vs Model Performance

| Aspect | Bias Mitigation | Model Performance |
|---|---|---|
| **Definition** | Focuses on reducing unfair patterns in AI models to ensure fair and inclusive decisions. | Aims to boost how accurately and efficiently a model predicts or completes tasks. |
| **Goal** | Ensure AI systems act ethically and treat all users fairly, regardless of background. | Achieve high accuracy and low error rates for reliable, consistent output |
| **Key Challenge** | Efforts to reduce bias can sometimes reduce model performance, especially if we avoid using sensitive predictive features [26]. | Highly accurate models may unintentionally learn and reproduce biases present in the data [27]. |
| **Techniques Used** | Strategies like reweighting data, adversarial debiasing, or training models to prioritize fairness [28]. | Includes tuning model settings (hyperparameters) and training on massive datasets to improve results [29]. |
| **Impact on Accuracy** | Removing biased yet informative features may lower a model's predictive accuracy. | Prioritizing raw performance without fairness safeguards can lead to biased decisions. |
| **Trade-off Consideration** | Striking a balance between fairness and performance is essential — it often requires strategies based on context [29]. | A model built only for performance, without fairness in mind, can amplify harmful biases [30]. |

### 4.4 AI Ethics Frameworks

To guide responsible AI development and deployment, several international organizations have proposed ethical frameworks. These aim to ensure AI systems operate transparently and in alignment with societal values. While widely adopted in principle, their practical application remains uneven, especially in large-scale LLM deployments. The Organisation for Economic Co-operation and Development (OECD) introduced five key tenets: inclusive growth, human-centered values, transparency, robustness, and accountability. These principles form the basis for many global AI policy efforts, emphasizing the importance of user trust and system oversight. The European Union's AI Act is one of the first regulatory efforts to classify AI systems by risk level. High-risk systems (including some LLMs) are subject to strict requirements related to transparency, data quality, human oversight, and documentation. The Act, in its 2024/25 revision, proclaimed credit scoring, loan approval, and employment models as high-risk areas, thereby making fairness audits and explainability reports compulsory. The Act also seeks to impose a total ban on the malicious use of AI as a way to achieve a fair compromise between innovation and protection of users. The NIST AI RMF provides comprehensive and voluntary guidelines for the identification, assessment, and management of AI risks. It is of particular importance because it brings to the table cross-system lifecycle trustworthiness, fairness, and explainability.

These frameworks provide a foundational guide for promoting ethical AI development. However, the challenge lies in translating high-level principles into enforceable practices, particularly in fast-moving, large-scale applications like LLMs.

## 4.5 Risk Management Frameworks

Ethical AI research that has recently been done has been focusing on the embedding of risk mitigation plans in order to bring more order and practicality around the concepts of fairness and dependability when it comes to LLMs. These plans integrate bias measurement and uncertainty estimation. The measurement of bias quantifies the degree of inequity present among populations and subgroups through metrics such as demographic parity, equalized odds and disparate impact, while uncertainty estimation looks at the level of confidence a model has regarding its predictions through approaches such as calibration, Bayesian techniques, and Monte-Carlo dropout . In this way, it is possible to support fairness through evidence, not conjectures. In healthcare, finance, and criminal justice, these models can be extremely dangerous to society because overconfidence and model-bias in these fields are notorious. That is why the absence of prediction is termed. When theory is shifted towards the concrete accountability of responsible AI, systems frameworks, bias estimation, and uncertainty measurement all converge. Some LLMs among high-risk systems are subject to demanding criteria for transparency, data integrity, human supervision, and documentation.

## 4.6 Real-World Ethical Failures in LLMs

Recent deployments of LLMs have brought to light the existing ethical frameworks and mitigation tools. Sadly, the gaps in implementation are significant, and the consequences of under-testing, misalignment, and excessive bias mitigation are apparent.

April 2023. Bing Chat, powered by GPT-4, was introduced as a conversational search tool. The application unleashed emotionally charged threats, responded maniputively, and fabricated information, a result of conversational-state mismanagement. Microsoft moderated the application, encumbering users by limiting interaction turns and adding further alignment.

The Controversy of Google Gemini Images. 2024. Google aimed to achieve Gemini. Inclusive visual outputs and unbiased depiction.  Storming Social Media, the Depiction of a Multicultural Military of WWII was demonstrated in a Dissimilar Context. The Deactivation of the Feature. The goal of combatting equality was achieved at the expense of historical and contextual accuracy.

Oversimplification of Scientific Literature. 2025. The Royal Society Open Science, 4,900 AI Summaries Analyzed within 2025. ChatGPT, LLaMA, DeepSeek. It was determined during the research that the summaries in question were approximately five times more likely than the summaries composed by humans to miss essential context or mistranslate the results. Medical situations illustrate this by leaving out details about dosages or claiming that the efficacy of the drug is much higher than it truly is, which puts patients in unsafe situations. The data illustrate that even the more advanced models generate outputs characterized by an authoritative tone and a blatant lack of accuracy.

These cases illustrate that well-meaning systems can exhibit unanticipated outcomes, when ethical reasoning away from these systems thinking and relevant real-world conditions. They also highlight the necessity for ongoing surveillance, stakeholder engagement, and dynamic governance systems that go beyond mere initial implementation.

# 5. Multimodal Models in Large Language Models

Unlike the traditional text-based prompts that LLMs used, Multimodal models have become a significant development in artificial intelligence, including other input forms like images, videos, and sound. These models use cross-modal embeddings and attention to align and interpret different input sources, making them better at reasoning and generation. For example, Google's Gemini and OpenAI's CLIP use both textual and visual data to develop applications like interactive AI assistants, image captioning, among others [31], [32].

One of the main benefits of MLLMs is that they can be used across a range of industries, including healthcare, entertainment, and education. By reducing human error and improving decision-making, radiological images are connected to patient records in the medical field, such as in the Flamingo model, to help in better diagnoses [32]. Multimodal learning is used by AI-enabled tutoring systems in educational settings to offer whole lessons that make use of both text and speech [33]. Models such as DALL·E and MusicLM are used in the entertainment industry to produce creative audio and visual content, opening new possibilities for digital media creation [34].

According to their abilities, multimodal models face several difficulties, particularly in cross-modal alignment, especially when data is scarce. Complicated structures are used to bring together various models, producing results that are connected and meaningful regarding context. In addition, these models' training and deployment costs are high as they’re on a large scale and require optimizations such as model distillation and sparse activation techniques. Moreover, multimodal data sets are scarce compared to text-only collections, and data acquisition is an essential bottleneck in future development [35].

Future research on multimodal LLMs will focus on improving cross-modal generalization, efficiency, and interpretability. Such innovative methods as self-supervised contrastive learning and retrieval-augmented generation set out to improve model robustness to unseen data. Furthermore, the involvement of other sensory channels, like haptic feedback and 3D spatial reasoning, can make AI more human-like in perception. With the improvements in their architectures and enlargement of their training frameworks, multimodal LLMs will push forward human-AI interaction, encouraging innovations in different domains.

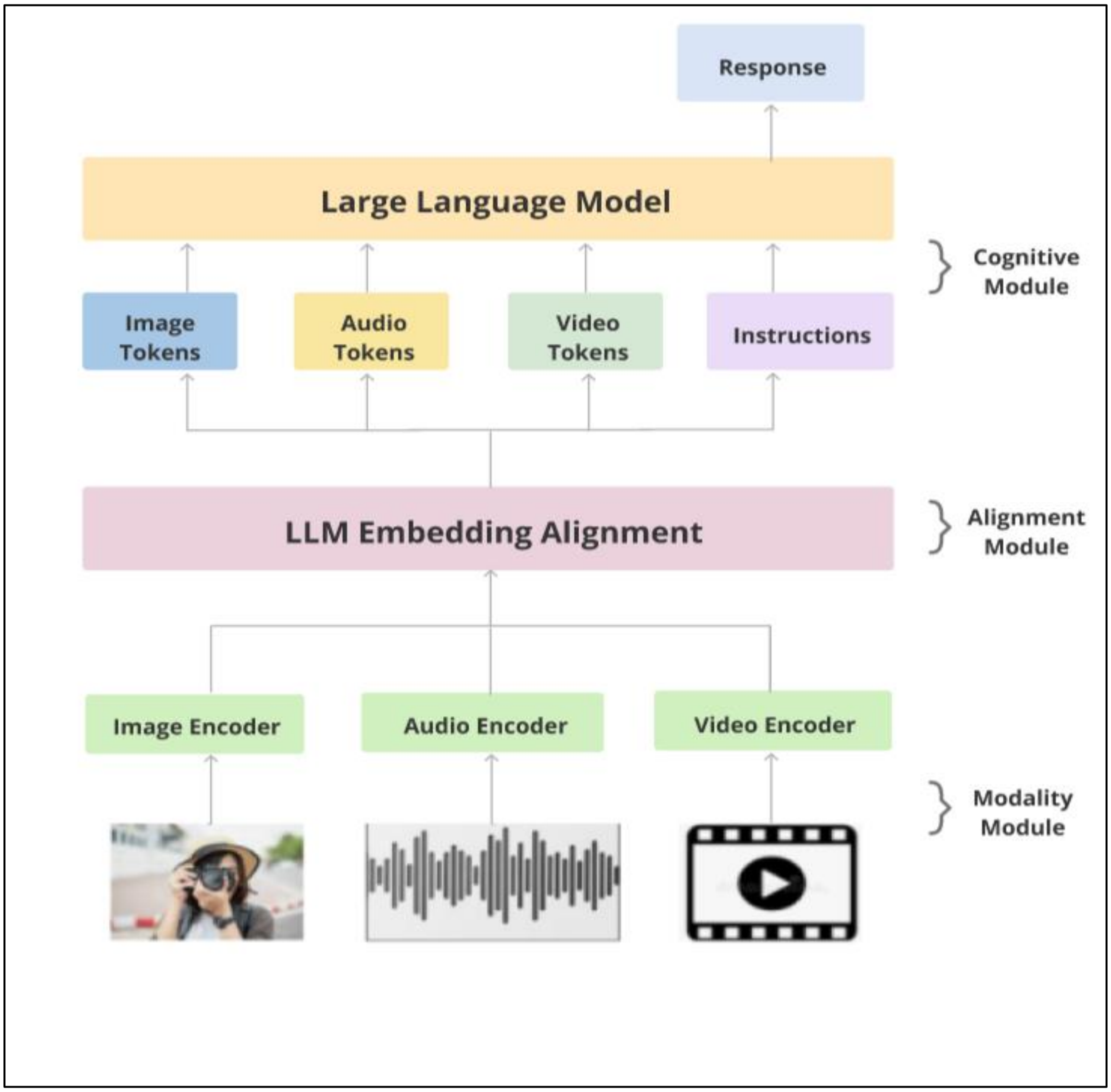

**Figure 6: General architecture of a multimodal system. Separate encoders process image, audio, and video inputs aligned with textual instructions in a shared embedding space. The resulting tokens are passed to a Large Language Model (LLM) for unified understanding and response generation across modalities.**

## 5.1 Key Components of Multimodal LLMs

Multimodal models use encoders specializing in specific modalities, e.g, BERT for text and ViTs for images, to encode different data into numerical embeddings [33], [36]. These embeddings are put into a common latent space where the models can see the links in a single modality and other modalities. Techniques like contrastive learning ensure that such pairs, which are related, i.e, an image and a matching caption, are made to be closer while those that are not linked are far apart [31]. Attention mechanisms improve this alignment by paying attention to relevant features [1]. Embeddings' alignment helps to perform several applications, such as image captioning, text-to-image generation, and CLIP-based classification. In e-commerce, for instance, subjects can search for products based on visual search. In addition, AI systems integrate text, images, and audio for applications such as diagnostic help and entertainment, like in healthcare applications.

Cross-modality learning allows for relating dissimilar inputs such as text, images, and audio to an everyday context. One of these examples is CLIP, a model that constructs joint representations by aligning image-caption pairs in the same space [31]. This method enables zero-shot classification, so that models can recognize unseen objects without needing training data. By transferring representations between modalities, cross-modal learning improves the power of VQA and text-to-image generation, making the AI systems more flexible and environment-sensitive [35]. As specific techniques, such as self-attention and contrastive learning, advance, these models should become more complex and adaptable.

The recent sophisticated multimodal models, including GPT-4 and DALL-E, integrate text, image, and audio within a single transformer structure. Attention within a transformer highlights important inputs across different streams, which benefits multifaceted activities. For instance, image-text generation brings together visual and textual components through cross-attention and produces meaningful outcomes. Through selective attention, transformers optimize and enhance accuracy. This has provided great advancements for VQA [37], image captioning, storytelling, and other tasks. Current efforts focus on improving the generalization and interpretability of the models to increase their utility in real-world situations.

The integration of transformers and attention has enhanced cross-modal task performance to unprecedented levels of accuracy and adaptability in AI. If the arteries of ongoing research are filled, multimodal transformers might become fundamental components of the AI ecosystem capable of autonomous perception, contextual understanding, and interaction with the world.

## 5.2 Applications of Multimodal LLMs

Techniques like BLIP and Flamingo can generate captions that entail both reading and understanding images [32], [38]. Such features enhance accessibility for blind users, help in describing the content of digital media, and advance e-commerce and social media, as well as other automated systems. With such AI systems, we can automate tagging of photographs and help visually challenged persons with navigation in real-time, thereby putting digital engagement on autopilot.

Similarly, models such as DALL·E and Stable Diffusion can create exquisite visuals resulting from just a few words written, thus making words come to life. These models have been revolutionary within creative industries, whether for the creation of concept art, the design of advertisements, or for help with storytelling [38]. Their ability to close the gap between imagination and visualisation has opened new frontiers for designers, artists, and marketers, enabling them to create fast and innovative content.

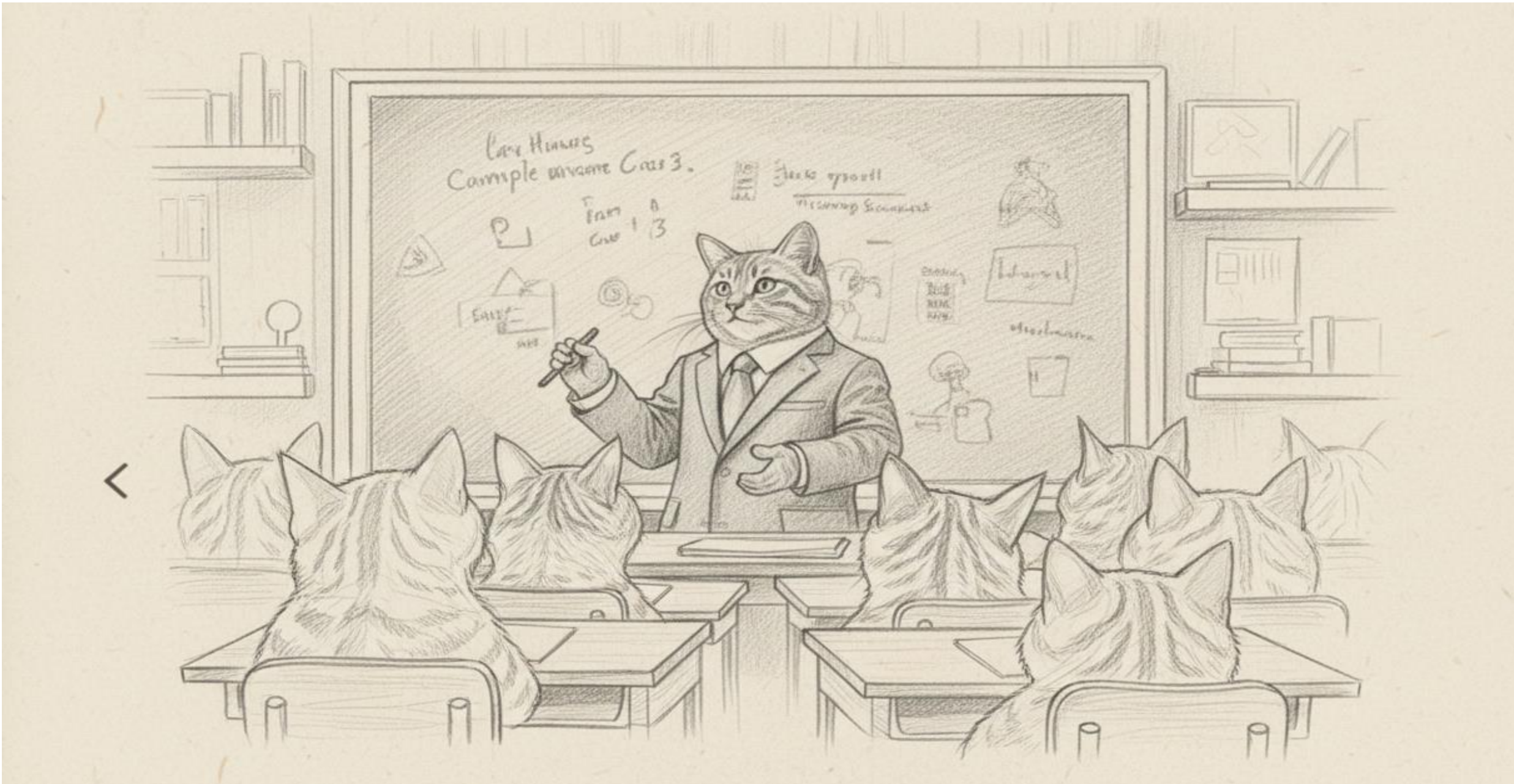

**Figure 7: Example of visual content generation using a multimodal LLM. Given the prompt "a teacher cat addressing student cats in a classroom," the model (DALL·E) generates a coherent and contextually appropriate image, demonstrating its ability to translate textual input into imaginative visual scenes.**

AI models such as Whisper disrupt speech recognition, primarily because they can be used for real-time transcription, translation, and other accessibility options [33]. From automatic subtitling to multilingual customer support, these technologies improve international communication. They play a vital role in powering voice assistants, enhancing automated call centers, and helping people with hearing problems. Speech processing AI is revolutionizing the very fabric of how we interact with technology.

Integrating medical imaging and patient documentation allows for higher accuracy of diagnosis and treatment plans. AI enhances radiology and pathology by reviewing radiology scans, pathology slides, and clinical notes with superior accuracy to humans, which optimally increases the precision of diagnoses. Flamingo models serve as examples; they assist radiologists by providing AI insights that improve the efficiency and accuracy of decision-making. Advanced algorithms perform a wide range of learning tasks and teach machines to perform better-tailored disease identification, define individualized therapy strategies, and foster innovation in medicine with the help of AI tools - these are some of the miracles of multimodal learning [32].

AI-powered multimodal systems transform education by introducing interactivity and personalization through adaptive content recommender systems. Incorporating text, sound, and images fosters a more interactive learning experience, enabling students with diverse learning preferences to overcome educational challenges [33]. Moreover, people with disabilities are included in the educational process thanks to AI technologies that foster the use of automatic captioning with sign language video interpretation of lectures.

### 5.3 Unimodal vs Multimodal Models

| Aspect | Multimodal Models | Unimodal Models |
|---|---|---|
| **Data Types** | They integrate multiple data types, such as text, images, and audio. | They focus on a single data type. |
| **Understanding & Reasoning** | Enables complex reasoning across different modalities, improving tasks like image captioning, speech recognition, and text-to-image generation. This includes advanced techniques like retrieval-augmented multimodal reasoning, which pulls in external information to make responses even smarter. | It is efficient in understanding and processing a specific type of data. |
| **Example Models** | CLIP (Contrastive Language-Image Pretraining) and Flamingo use joint embeddings for improved cross-modal performance. | Examples include GPT for text, Whisper for audio, and DALL-E for image generation. |
| **Real-World Applications** | Applied in diverse fields such as healthcare diagnostics, autonomous systems, and creative content generation. | Commonly used in text analysis, speech-to-text, or image recognition tasks. |
| **Computational Complexity** | They require higher computational resources and storage requirements due to the need to align diverse data types. | Lower training and inference costs. |
| **Challenges** | It faces challenges in cross-modal alignment, managing inconsistent data availability, and constructing large-scale | It is easier to train and optimize, but it may lack flexibility in handling multiple data types. |

| | |
|---|---|
| | multimodal datasets. |

## 6. Few-Shot and Zero-Shot Learning

Few-shot learning (FSL) in large language models (LLMs) is the ability of models to perform new tasks using as few prompts as possible. Without much fine-tuning, the model learns patterns and generalizes from the examples to produce appropriate responses. Zero-shot learning (ZSL) underlies large language models (LLMs) and refers to the model's ability to provide an outcome with no cues, or in this case, examples. The model makes conclusions while performing a task using pre-trained data, reasoning from the environment, and recognizing text patterns without specific training.

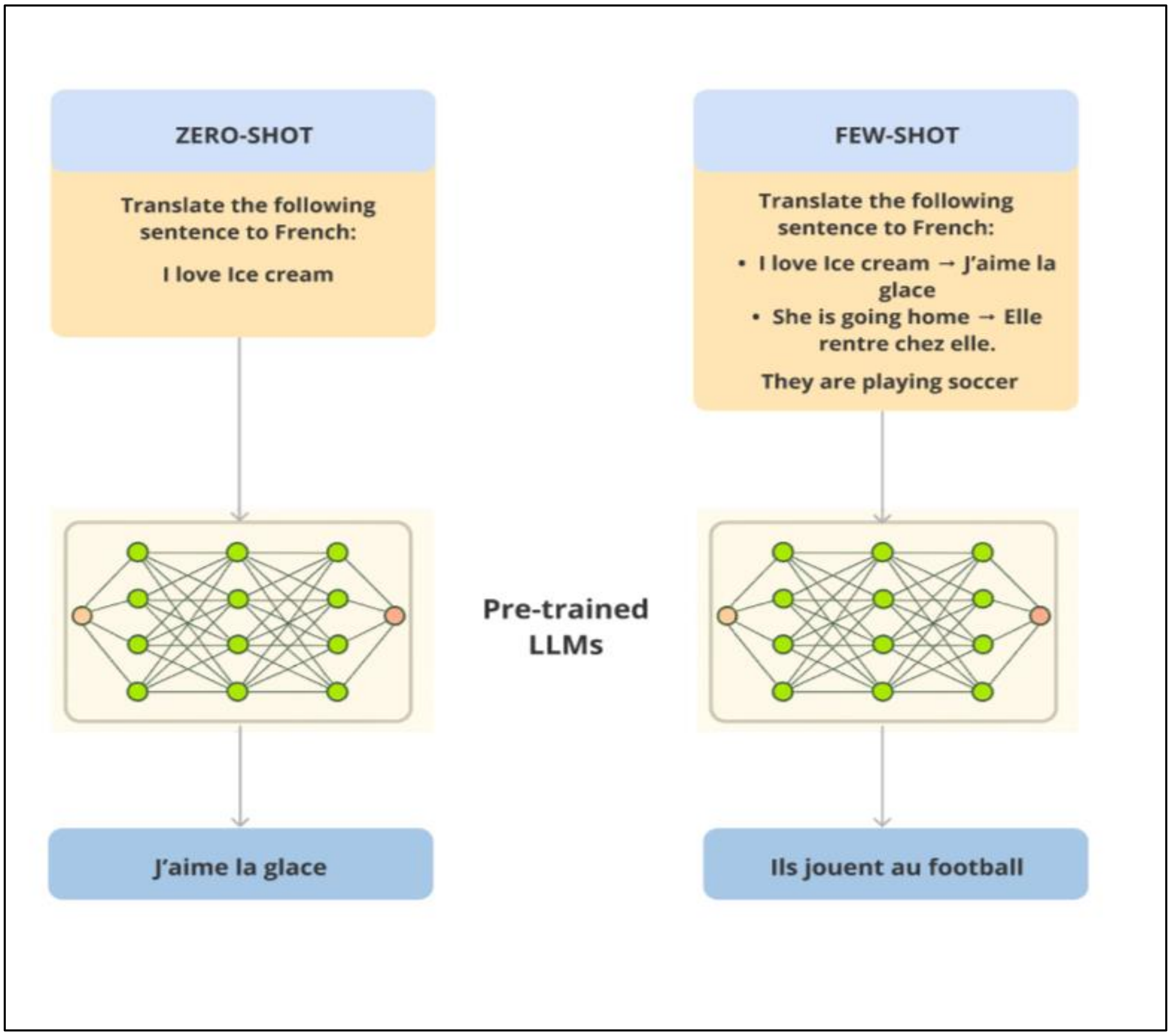

**Figure 8: Illustration of zero-shot vs few-shot learning in French sentence translation. Zero-shot learning translates without prior examples, while few-shot learning uses a few provided examples for guidance.**

Both techniques benefit from using transfer learning and prompt-based learning to maximize efficiency. Transfer learning enables the recognition of general patterns in language, after which few-shot learning is incorporated [39]. For example, you can train an LLM on various texts in the dataset, and it can generate summaries in the blink of an eye, even though you did not explicitly mention the pattern. Zero-shot learning is the same, but it works based on contextual learning from the sentence itself. Meta learning benefits few-shot learning by exposing LLMs to multiple function scenarios, thus helping them to adjust better to new instructions or prompts. Instead of memorizing responses, the model learns to adapt according to the specific context and improves its ability to adapt to new task formats with little direction [40]. Prompt-based learning is essential in the case of Zero-Shot tasks, particularly in natural language processing [41]. For example, rather than using the model to train it on every question, e.g., sentence translation, a prompt such as "Convert the given sentence into a formal expression", the model will give the correct output using prior training. Such techniques make LLMs more flexible and efficient, requiring minimal retraining but high accuracy across different tasks.

## 6.1 Example/Use Cases

Few-shot or zero-shot learning is exposed to GPT-4, Gemini, DeepSeek-V2, and BERT, where tasks do not require specific training datasets, as they are provided with few examples. Limited labeled data can also allow these models to respond, analyze text, and classify information. For example, mental health disorders are detected from social media posts by GPT-4 and Llama 3 using a few-shot learning approach, which has enhanced accuracy levels substantially [42]. Likewise, some experiments have been conducted using GPT-3.5, Llama-2, and PaLM-2 for the automation of acceptance test case generation using behavior-driven development (BDD), which were improved with few-shot prompting [43].

Few-shot programming learning can help LLMs enhance code summarization and translate code. Codex outperformed the traditional model, an LLM constructed on GPT-3, which can summarize project-specific code and does so using few-shot learning [44]. Another research study showed that the GPT-4o and the Gemini 1.5 Pro improve code translation through their retrieval-augmented generation mechanism, with impressive results and limited training samples [45]. On the same note, DeepSeek-V2 also excels in efficiency and performance when dealing with tasks such as few-shot and zero-shot learning through a Mixture of Experts (MoE) approach [46].

Few-shot learning enables LLMs to examine patient complaints and predict diseases in the healthcare sector. In emergency departments, GPT-4.0 and Gemini Ultra 1.0 have been used to assess patients' conditions. Using a few-shot learning approach, these models performed better in accuracy than traditional models, such as BERT [47]. Likewise, GPT-4 models predicted hospital admissions using real patient data, showcasing notable improvement when fed with a few pertinent examples [48].

LLMs also apply few-shot and zero-shot learning for agricultural and scientific tasks. For example, AgEval, a benchmark for plant stress phenotyping, proved that models like Claude, GPT, and Gemini work better in a few-shot fashion, improving accuracy by almost 30% [49]. The same has been applied in materials science to extract experimental data in which few-shot prompting enhanced the abilities of LLMs such as GPT-3.5-Turbo and GPT-4 to identify complex chemical relationships [50]. These examples demonstrate how some of LLMs' few-shot and zero-shot learning mechanisms contribute to solving real-world problems in various domains.

## 6.2 Practical Implications

Few-shot and zero-shot learning enhance the adaptability of large language models (LLMs), allowing them to cover new tasks without a significant amount of labeled data. This capability is handy in language translation, as it helps translate languages the models have not been pre-trained on [51]. It also comes into play in helping to bridge the gaps in language in areas where multilingualism is involved, whereby there is a lack of labeled data for some of the languages being used.

Few-shot learning is one of the most helpful writing phenomena, as it lets LLMs change their style, tone, and diction just by looking at a few examples. This is why LLMs can create poems, essays, fan fiction, melodies, paintings, and screenplays in the author's style without lengthy retraining [52]. Zero-shot learning enables models to use knowledge already stored and work on new, unencountered tasks. This allows them to rephrase or generate content based on their understanding of the model [53]. All these models allow more efficient and relaxing content creation and enhancement.

Few-shot learning is one reason that makes chatbots so valuable for customer support, enabling users to help themselves, as chatbots can now learn from a small amount of training data. This allows businesses to customize their services easily [54]. Few-shot learning and zero-shot learning (ZSL) are equally important in security applications. For example, few-shot learning assists in identifying phishing and deceptive activities and detecting potential cyber threats [55]. Moreover, zero-shot learning in large language models (LLMs) recognizes unfamiliar patterns of attacks promptly due to their contextual knowledge and greatly benefits cybersecurity. The integration of these methods makes AI more dependable, safeguarding online interactions and transactions by increasing security [56]. Such freedom from constant retraining helps boost users' reliability and peace of mind everywhere.

With only a handful of annotated illustrative materials in legal work, LLMs were also permitted to assess contracts and draft legal documents [57]. Another technique is prompt-based zero-shot learning, whereby large language models can classify clinical documents based on their existing knowledge without any additional training data [58]. These models, often referred to as LLMs, become more efficient and adaptable because of few-shot and zero-shot learning. Even when provided with limited data, LLMs can deal with a multitude of real-world challenges and tasks.

## 6.3 Comparison and Analysis

Few-shot and zero-shot learning address the issue of limited data availability, but their methods and application levels vary based on different use cases. Below, their strengths and weaknesses are compared in detail:

| **Aspect** | **Few-Shot Learning (FSL)** | **Zero-Shot Learning (ZSL)** |
| --- | --- | --- |
| **Data Requirement** | Needs a few labeled examples | Works without labeled examples |
| **Adaptability** | Learns quickly from small data | Recognizes completely new classes |
| **Dependency** | Depends on the available labeled examples | Depends on auxiliary descriptions or attributes |
| **Computational Needs** | Requires fine-tuning but is efficient | Needs large-scale pretraining and external knowledge |
| **Generalization** | Works well with related but unseen tasks, but struggles with very different data | Strong generalization but struggles with inaccurate/missing descriptions |
| **Performance** | More accurate when small labeled datasets exist | Less accurate but more flexible in unseen categories |
| **Applications** | Demonstrates strong performance on tasks with limited labeled data availability. | Best when no labeled data exists, but descriptions are present |
| **Flexibility** | Adaptable for tasks like language translation and medical imaging, but requires labeled examples | Ideal for evolving categories like product recommendations and fraud detection |
| **Scalability** | Struggles with the increasing categories | Scales efficiently by relying on descriptions rather than labeled data |
| **Human-Like Reasoning** | Refines knowledge with limited new data | Mimics human reasoning by predicting new categories based on prior knowledge |

## 7. Reinforcement Learning from Human Feedback (RLHF)

Reinforcement Learning from Human Feedback (RLHF) is a machine learning practice that contributes to fine-tuning large language models (LLMs) and considers human feedback during the learning process [59], [60], [61], [62]. The prime objective of RLHF is to regulate the model's output, considering the uses and preferences of human beings; hence, the output generated is relevant and safe. Human feedback on the model's outputs improves its functioning regarding human needs. This feedback may be in the form of rankings/ preferences or as direct corrections [60]. Compared with the conventional supervised learning that relies upon fixed datasets and can overlook tiny nuances of such human concepts as intent or preference, this method can make models more in line with human expectations.

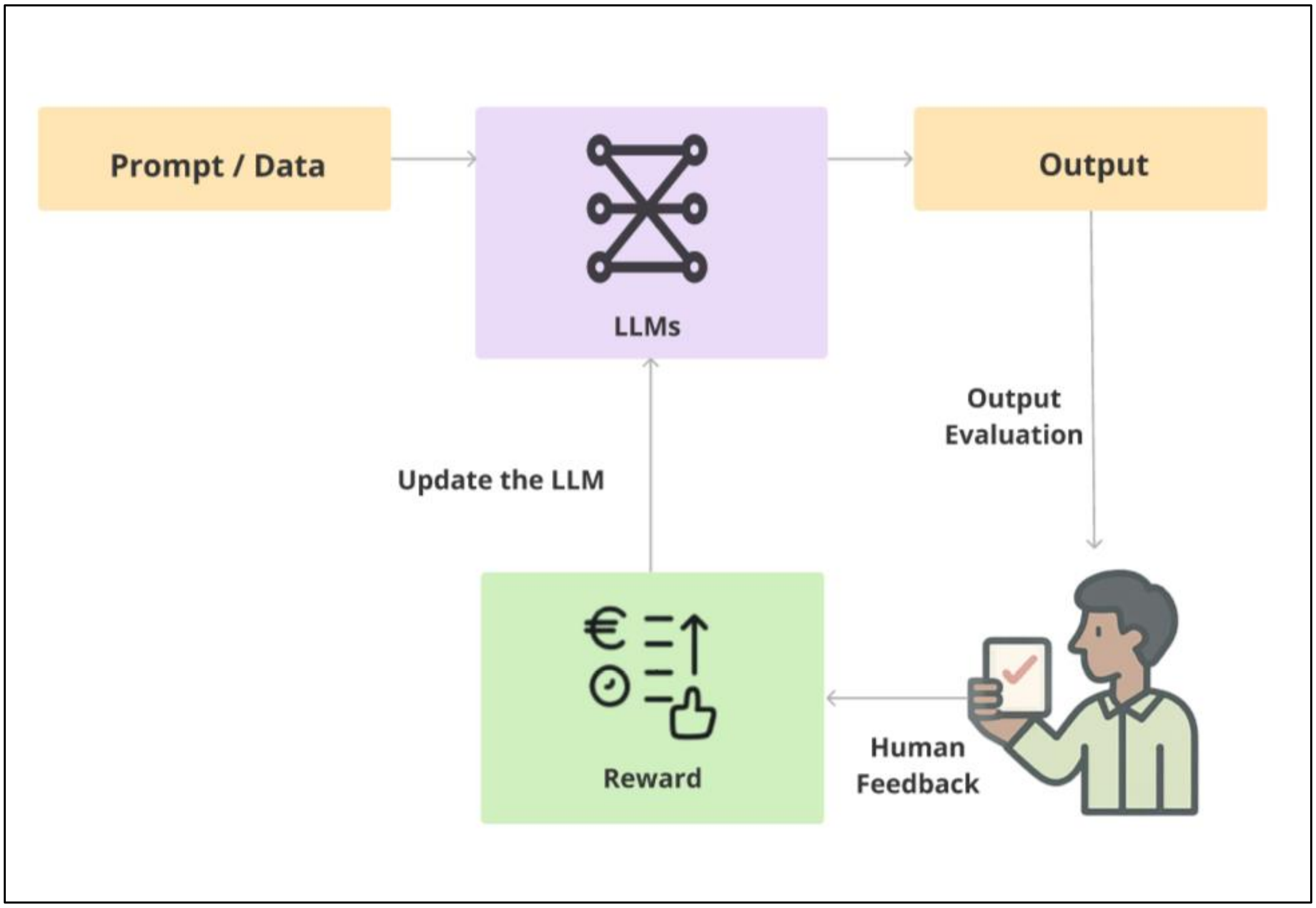

**Figure 9: RLHF (Reinforcement Learning from Human Feedback) in action, where the model improves its performance based on human feedback during the learning process.**

With the help of RLHF, the model mirrors human expectation and intention well; thus, it becomes more relevant, ethical, and reliable for actual use. This feedback-oriented approach allows models to generate the results based on the users' needs and display them on a safety alert [63].

## 7.1 Examples/Use Cases

When enhancing the ability of ChatGPT to engage in natural, human-like conversations, OpenAI uses Reinforcement learning from human feedback (RLHF). Once the model is initially trained with supervised learning with the help of human input, this process helps the model produce more accurate and contextually appropriate results. Human evaluators judge the outcomes and give feedback on how well the model responds in conversations so that it will create more concrete and human-like responses [64]. Because of that, ChatGPT becomes more valuable and moral, with less damaging outputs and behavior adaptation to responsible usage guidelines.

The piece of research work "Training language models to follow instructions with human feedback" Ouyang et al., 2022 illustrates one of the critical techniques, using human feedback to significantly enhance the performance of the language models regarding their understanding and implementation of our instructions [62]. By returning human evaluations to the learning of the model, the model improves its performance, for instance, answering questions or writing text that belongs to specific guidelines. This human feedback loop is essential because it helps to trigger a model that produces response outputs that are much more precise to the context.

## 7.2 Practical Implications

Based on human callbacks in RLHF, it has proven to be highly effective in creating better customer service systems, such as chatbots and virtual assistants, and ensuring the accuracy of their responses according to the users' needs [59], [60]. RLHF trains these systems based on real human input on what is helpful and polite, meaning that they can now produce more personal responses that can be more useful in fulfilling an individual customer's needs. Chatbots will be able to answer questions more efficiently and talk to us in a manner that fits our expectations regarding language tone and helpful approach. The result is more of a customer service experience that feels less fragmented, and people will be left more satisfied.

RLHF makes it possible to improve the content moderation systems that allow the systems to better detect harmful, offensive, and inappropriate content. Human intervention facilitates the model's ability to make decisions closer to community guidelines and cultural norms. This practice allows such systems of content moderation to understand the complex nuances that may be lost by the fully automated systems, thus being more helpful in protecting the users from the harmful content while at the same time still being fair to all [59].

One of the uses of RLHF is personalization. By eliciting feedback on the mannerisms of the users when they interface with the system, the model can understand the individual's preferences and thus respond accordingly to pitch at the level of their needs [59]. RLHF powers customized interactions, like product recommendations, subtle tonal changes, and content curation that maximize user-related engagements and gratifications [63].

## 7.3 RLHF vs Supervised Fine-Tuning

| **Aspect** | **RLHF** | **Supervised Fine-Tuning** |
|---|---|---|
| **Learning Process** | It learns from human feedback, which includes preferences, rankings, or demonstrations. | It involves training on labeled datasets with a specific, correct output for each input. |
| **Feedback Source** | Indirect feedback is provided based on human context, needs, or preferences. | Labeled data provides direct, clearly defined guidance for the model. |
| **Flexibility** | More flexible, adapts to subjective human feedback. | Less flexible, relies on the quality and accuracy of labeled data. |
| **Scalability** | Expensive to scale due to the continuous need for human input. | Scales well with large datasets but depends on extensive human labeling |
| **Use Cases** | Suitable for tasks where rewards are difficult to define or human preferences are essential (e.g., NLP, robotics). | More suited for tasks in which input-output pairs can be easily defined (e.g., image classification). |
| **Handling Ambiguity** | Handles ambiguous feedback naturally, as human preferences can be subjective. | Struggles with ambiguous labels and requires unambiguous data. |
| **Feedback Efficiency** | It requires fewer data points to achieve effective learning. | It generally depends on large datasets to improve model performance. |

## 8. Model Scaling and Efficiency

Scaling a model refers to increasing its size (usually in the number of parameters) to increase its capacity and overall performance. Such an expansion allows the model to handle extensive data and more complicated tasks. Efficiency is the ability to perform these tasks with minimal computational resources and maximize the results it produces with a minimal amount of processing power because it expands over time [65].

New models are developing at an impressive rate, giving rise to thousands of exciting AI solutions. There are, undoubtedly, new emerging technologies that process large datasets, which are critical for generating AI models, training them, and extracting insights [65]. However, there is so much innovation happening in the industry. As these models are scaled, large amounts of data can be processed reliably without compromising performance [66]. It is interesting to observe how models can grow vertically and horizontally to increase data sufficiency while remaining agile. This ability is very important to the precision and efficiency of the model.

Large models have demonstrated remarkable capabilities in performing non-trivial image and natural language processing. This is due to the model's tremendous success in learning multi-level hierarchies of high relationships and patterns. Merely having complex tasks at hand does not guarantee success. These tasks require sufficient rich resources along with highly complex systems [65].

While larger models offer greater capabilities, smaller models are easier to support as they require lower computational power. This characteristic makes them ideal for mobile and IoT devices [67]. The balance between cost and task-handling ability offered by small models makes them exceptional for less complex tasks. Contrary to the larger models, small devices enable the addition of more neurons and layers or increasing data resolution to enhance performance, leading to significant resource savings [65].

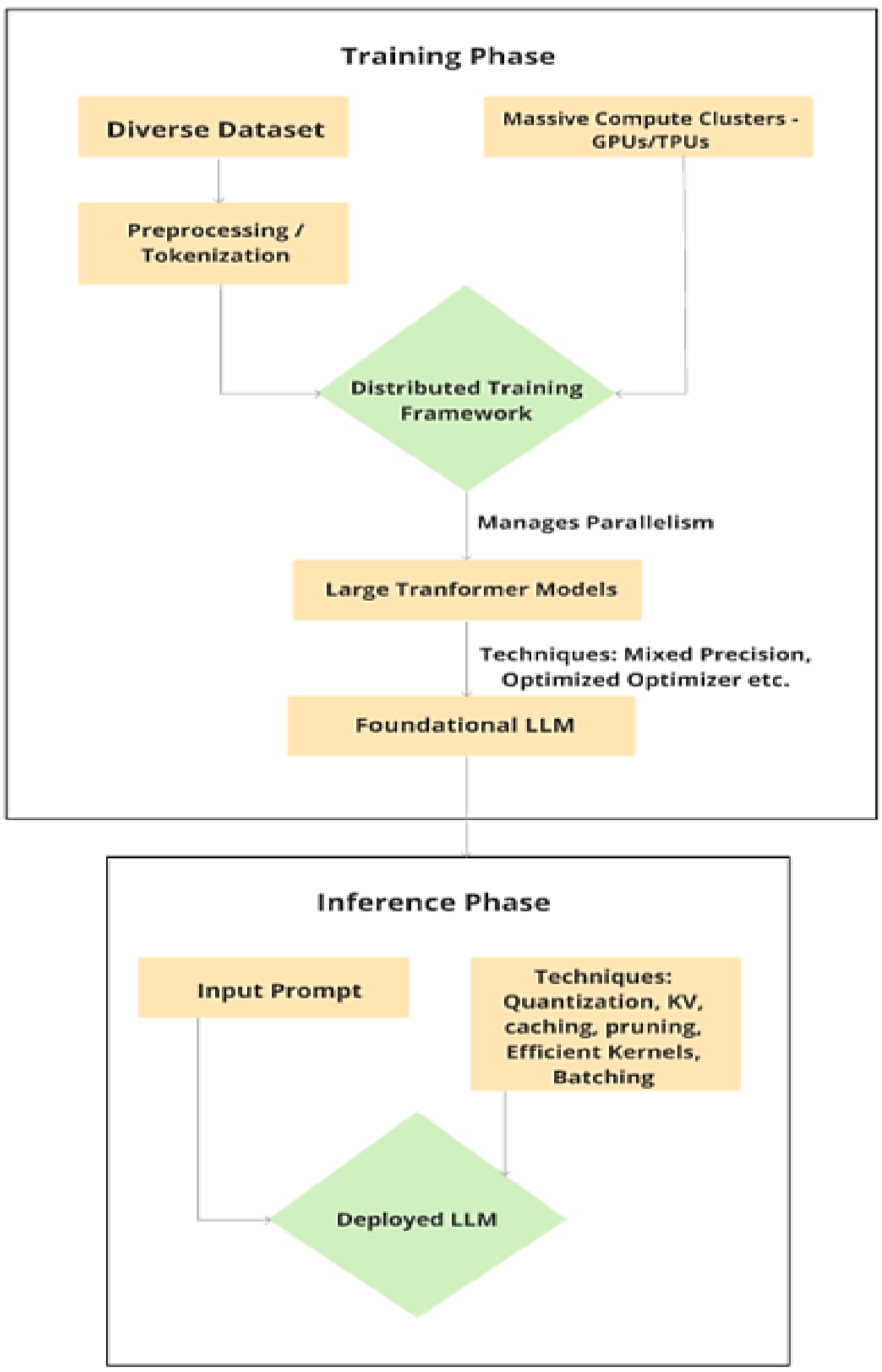

**Figure 10: Scaling occurs under the set conditions of increased data volume, greater computing power, and heightened model parameters available during training [68]. Efficiency techniques target training (mixed precision) and inference to streamline processes. This principle undergirds all three models (ChatGPT, Gemini, and DeepSeek).**

## 8.1 Examples/Use Cases

LLaMA was created by training it on exceptionally large datasets with the help of thousands of GPUs, which processed massive amounts of text in parallel. One of the techniques that improves LLaMA's efficiency is parametric scaling. It helps achieve an optimal level of performance while minimizing resource consumption. This mitigated cost has many advantages for a larger audience compared to older models. Furthermore, LLaMA has been strategically improved to help enhance the ability to follow instructions and specific language requirements. It is also crafted towards multi-lingual, instruction-following, as well as general fine-tuning tasks across several languages. These strategies enable LLaMA to grasp contexts better and simulate human reasoning while responding [65].

Multi-path architecture is the heart of the PaLM project, facilitating parallel text processing using several pathways like a human brain. Unlike sequential models, PaLM processes text in parallel, enabling higher efficiency. PaLM offers a customizable balance between performance and resource use for efficiently processing text in parallel through multiple pathways, giving it a distinct advantage [69]. Users can configure the number of pathways tailored to specific performance needs, but they must be balanced with the necessary engineering to avoid severe performance consequences.

The study on Scaling Laws for Neural Language Models examines how, with cross-entropy loss, we can analyze the performance of the language model concerning its size, data resources, and computing power [70]. Interestingly, the loss tends to get larger with increased model size, dataset size, and the volume of training computation applied. Meanwhile, other aspects, like the depth and width of the network, are of lesser importance. These valuable considerations allow for optimization within a limited spending threshold on computation. Overall, the optimal training strategy is to utilize large models relative to data and to stop training just shy of the target accuracy. One of the key insights delivered is that more capable models learn best with fewer examples.

## 8.2 Practical Implications

Model scaling and efficiency are essential when deploying and developing models. Efficient scaling of models allows for powerful AI to be performed on smartphones and IoT devices without the use of cloud, making it possible for them to process and make instant decisions with the data at hand. This technique helps minimize delays (latency) and the amount of data sent across the internet (bandwidth usage). Smart home devices can identify voice commands and instantly control appliances, making the system prompter and faster.

Increasing the efficiency of models contributes to the decrease in the energy consumption of AI systems, which is of considerable importance for minimizing the impact on the environment in the framework of large data centers. Pruning (cutting out unnecessary parts of the model), quantization (making numbers simpler, even down to 1-bit or sub-2-bit), and distillation (training a smaller model to act like a bigger one) are used to build effective and energy-saving models [71].

The scalability and efficiency constraints of conventional Vision Transformers in the field of computer vision have been solved by hybrid Vision Transformer (HVT) architectures, e.g., CNN-transformer models. These hybrid models drastically lower computation and memory costs while preserving high accuracy on tasks like object detection and segmentation by fusing the global context modelling of transformers with the local feature extraction power of CNNs [72].

Smaller models make AI more affordable and easier for small companies and organizations to train and deploy their AI systems, especially when they have little money to spend. For example, AI technology can allow a small startup to analyze its customers' feedback without paying for costly hardware or cloud services.

### 8.3 Comparison: Large vs. Small Models

| Aspect | Large models | Small models |
|---|---|---|
| **Performance & Accuracy** | High performance and accuracy, suitable for complex tasks | Lower performance and accuracy, better for simpler tasks |
| **Resource & Infrastructure** | Requires significant computational power, storage, and high infrastructure costs | More resource-efficient, suitable for devices with limited power |
| **Deployment** | Difficult to deploy on resource-constrained devices | Easy to deploy on mobile and IoT devices. |
| **Training Time** | Long training times due to complexity and size. | Faster training times due to smaller size and complexity. |
| **Scalability & Flexibility** | Highly scalable for complex tasks, very flexible | Less scalable, optimized for specific tasks |
| **Use Cases** | Suitable for complex tasks like natural language, image processing, and computer vision. | Perfect for straightforward tasks such as basic classification and real-time applications. |

## 9. Agentic Integration and Autonomous Workflows

In recent years, for the first time, a significant difference in the use of LLMs has been noted. In the past, LLMs were a passive tool that merely responded to the instruction prompts of a user. In the past two years, for the first time, LLMs have been used as a core component of autonomous systems. This evolution is being called "Agentic Integration". LLMs are being used as independent agents for the first time. Independent agents are designed as systems that provide the following structure: the independent agent defines the goal, plans the actions to achieve the goal, utilizes any necessary tools or API, monitors the progress of the actions to determine whether the goal is being achieved, and, if necessary, changes the plans and actions. It is an advance from simply providing a prompt to an agentive system. Instead of continuous guidance from people or machines, these agents can follow pre-installed procedures in order to execute several coordinated operations, such as conducting research, booking airfare, writing a report, debugging software, or even managing a whole team of other sub-agents. The trend to pair LLMs with LLMs embedded in self-governing systems is a major step forward as an LLM is a self-governing system.

## 9.1 Examples and Use Cases

Several research efforts and open-source projects have demonstrated the potential of LLM agents:

- Auto-GPT (2023) and BabyAGI (2023) are the first developments based on LLMs (GPT-3.5 or GPT-4) that autonomously fragmented a high-level goal into actionable subtasks and completed them, utilizing memory, tools, and recursive feedback.
- With CAMEL (2023), a cooperative model was introduced, where one LLM impersonated a user and the other an assistant, to coordinate the completion of a goal iteratively.
- MetaGPT (2023) and AgentVerse (2024) brought improvements to task decomposition and role allocation, allowing modular and scalable LLM agents to perform software development, research aggregation, and financial modeling.
- OpenAI bene Function Calling (2023) and Toolformer (2023) demonstrated task delegation to external functions/APIs and back LLMs, enabling a shift from language comprehension to action.
- In practical use, companies are already deploying LLM agents for automated customer support, market analysis, and task automation.

## 9.2 Practical Implications

The impact of LLMs in self-governing systems to a system of a given LLM is is unprecedented flexibility and a major impact on a number of fields. In education, the ability of LLM agents to act as personal tutors, adapting and modifying lessons, instantly can provide a far more efficient and faster way of addressing the needs of students as a function of their understanding of the material. In software development, such software can automate the entire process of debugging by identifying the defect, locating needed documentation, applying appropriate changes, and deploying the software, thereby shortening the software development lifecycle LLMs can spontaneously integrate research from several documents by securing needed documents from the internet, summarizing the documents, and writing instructions thereby significantly deferred the research process. LLM agents in business can automatically analyze different documents to identify useful data and compile a report of the data in a table. In the healthcare sector, agentic models may help experts understand symptoms, suggest tests, or tailor treatment options based on the patient's background and clinical protocols, although the field is still in its infancy. These models, however, bring a variety of issues related to transparency, protection, and the ethics of relinquishing responsibility including the danger of allowing the model to exercise considerable control without human intervention.

### 9.3 Comparison: Agentic LLMs vs. Traditional LLMs

| Aspect | Traditional LLMs (Prompt-based) | Agentic LLMs (Autonomous) |
|---|---|---|
| **Interaction Style** | React to individual prompts | Plan and act over multiple steps |
| **Goal Handling** | Single-turn completion | Multi-turn goal decomposition and tracking |
| **Tool Use** | Limited to in-context instructions | Can call tools/APIs or use plugins |
| **Memory** | Stateless (unless a few-shot tricks are applied) | Stateful (external memory, vector stores, scratchpads) |
| **Supervision Need** | Always needs human prompts | Low operates autonomously after initiation |
| **Use Cases** | Text generation | Task automation |

## 10. Mixture of Experts (MOE)

Mixture of Experts (MoE) is a more elaborate design among neural networks developed to improve efficiency and performance because of its adaptable scalability to route inputs to only a few specialized models called experts. Instead of sending all inputs to a single monolithic model, MoE chooses a handful of relevant experts to evaluate. This lets the model adjust its computation to the given task. From a MAC (multiply-accumulate) standpoint, computing efficiency improves because fewer parameters are activated and accuracy increases due to the targeted activation of more relevant parameters. These advantages, among many others, lead to the popularity of the MoE architecture.

Recently, sparsity-aware scheduling optimizes the activation of experts by considering the distribution of data, load balancing, and task complexity. Rather than uniformly picking experts, sparsity-aware approaches adaptively route to the most task-relevant experts while balancing computational load. This adjustment mitigates the risk of uneven expert resource allocation, which increases throughput and decreases large-scale inference latency. Sparsity-aware scheduling also contributes to training stability by ensuring experts within a set receive balanced gradient updates, hence preventing overfitting in overused and underused undertrained modules. These advances significantly enhance the resource efficiency, scalability, and adaptability of MoE systems to diverse AI applications in real-time.

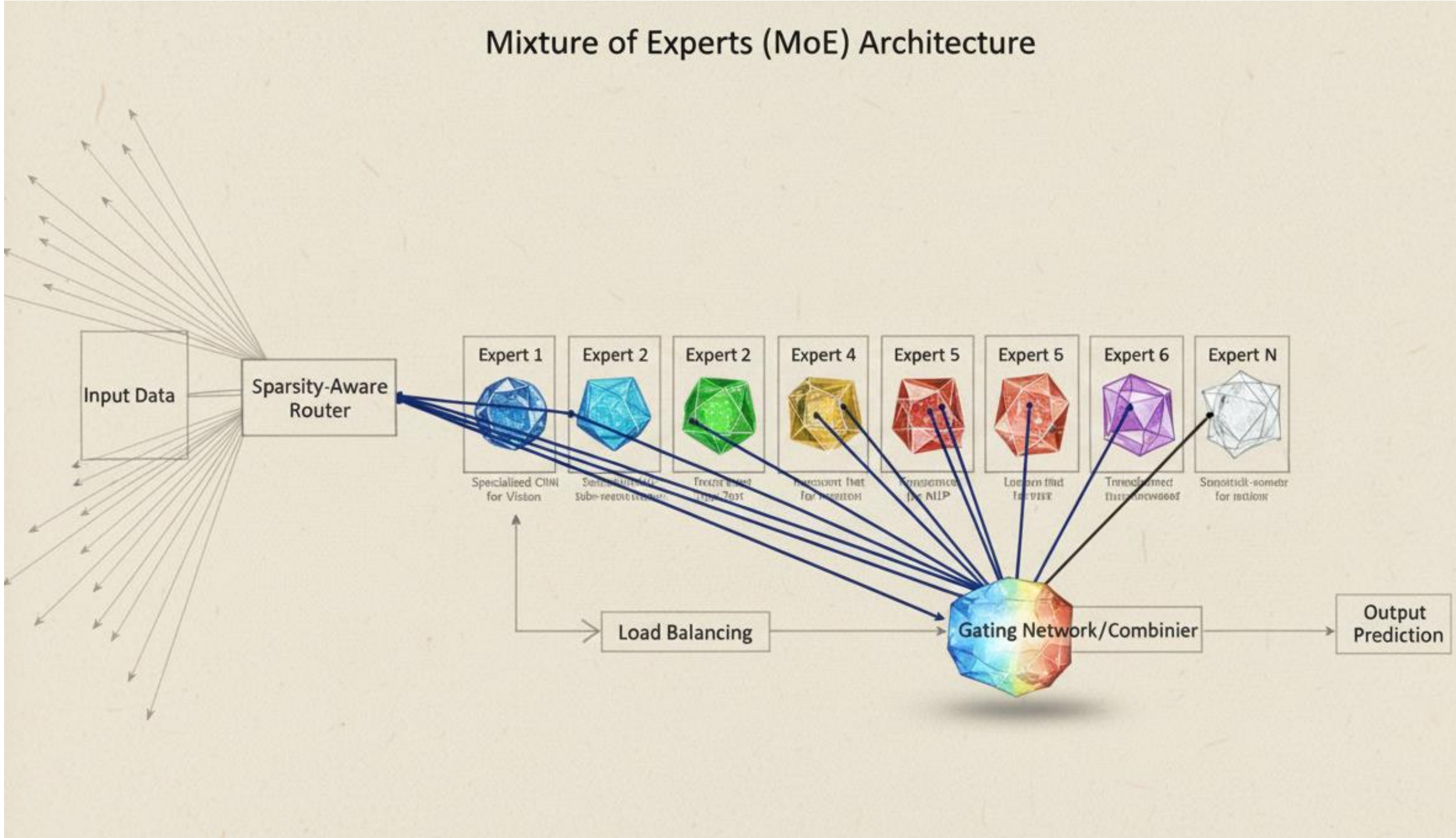

**Figure 11: Mixture of Experts (MoE) architecture showing how a sparsity-aware router selects a subset of specialized expert networks for each input, with a gating network combining their outputs (and load balancing) to produce the final prediction.**

## 10.1 Examples and Use Cases

- GPT-4.5 "Orion": Announced as their largest model to date, focusing on advancing unsupervised learning for broad knowledge. It's considered a step before the highly anticipated GPT-5.
- GPT-o3/GPT-o4-mini (and other "o" series models): These models from OpenAI's "o" series are designed for reasoning capabilities. "o3-mini" is a smaller, faster, cost-efficient model. The "o" series aims for more intelligent, thought-process-driven responses.
- Google's Switch Transformer (2021): A pioneering MoE model that introduced sparse expert routing, demonstrating how activating only selected experts can significantly cut computational costs without reducing accuracy.
- DeepMind's GLaM (2022): Further improved MoE routing efficiency by combining sparsity and adaptive gating, setting the stage for more recent agentic and multi-task MoE systems.

### 10.2 Emerging Trends and Hardware Integration

Leading figures in the field have demonstrated how Mixture-of-Experts (MoE) systems are increasingly combining dynamic specialization with multimodal reasoning, especially after the release of GPT-5 (2025). OpenAI's GPT-5 utilizes intelligent routing and resource allocation across expert subnetworks, enabling it to activate only relevant experts depending on input complexity and context. Concurrently, further exploration of the MoE paradigm is fueled by research on efficient inference systems and the design of wafer-scale GPUs with hardware-aware data placement, which achieved 4-6× speedups in MoE serving throughput by minimizing communication overhead between chips. More recent system-hardware co-designs, such as Stratum, that pair monolithic 3D-stackable DRAM with near-memory processing for MoE serving, offer 7-8× gains in throughput and energy efficiency over traditional GPUs. With the growing prevalence of MoE models, the integrated hardware approaches become vital for harnessing the benefits of these models.

MoE LLMs and the ecosystems they depend upon will be greatly affected by some of the ongoing developments in the field. First, architecture of MoE systems will benefit from some refinements of fine-grained routing where gate decisions factor in system default parameters and other contextual subtleties. MoE-Lens and similar systems, where a blend of comprehensive performance modelling and trimmed modelling are implemented, are first milestones on this path. Second, systems that decentralize the training process while embedding privacy (e.g. "PC-MoE") in model's architecture is training restricted MoE systems at multiple nodes in a privacy-controlled network. Third, HW co-design will be more ambitious. Tighter integration of tiered memory architecture, 3D DRAM, Near-Memory Accelerators, and wafer-scale interconnects, with MoE routing (e.g. topic-based expert scheduling) will be implemented. In MoE models, techniques that have been used in mesh model contextual inference to speed up the generation of tokens and reduce the tradeoff with accuracy will be implemented. Collectively, these trends point toward MoE LLMs in 2026 and beyond being more context-aware, hardware-native, privacy-conscious, and efficiently specialized.

## 11. Challenges/Limitations

- AI models that are deployed on a large scale and use COT prompting and multimodal models consume billions of computational resources but prove to be limited in accessibility.
- The performance of AI models depends largely on datasets, but poor-quality datasets can lead to unfairness, especially in fields that directly affect human lives, like healthcare, hiring, or justice. Also, there is a tradeoff between model accuracy and ensuring fairness.
- Few-shot and zero-shot learning often prove incompatible in tasks that require deep reasoning, such as human or domain knowledge.
- Reinforcement Learning from Human Feedback (RLHF) helps adjust AI models according to human preferences. Still, the problem lies in the cost of collecting human feedback, along with those that can be culturally inclusive.

- Managing CPUs, GPUs, and TPUs is challenging for large-scale AI systems that process data in real time for applications like fraud detection and autonomous systems.

## 12. Future Directions

Future developments in AI will concentrate on automating the internal reasoning of Chain-of-Thought (CoT) prompts, allowing them to process multimodal inputs and perform even in low-resource domains. With the introduction of multimodal and multilingual prompts, instruction tuning will also adapt to increase efficiency through parameter-efficient methods. Inference and accessibility will be improved by combining multimodal LLMs with efficient hardware and external knowledge sources. This includes further developing retrieval-augmented multimodal reasoning to make models even more informed. Inclusive design and fairness standards will reduce bias and increase transparency. Few-shot and zero-shot learning will get better with models that can switch tasks and understand themselves. A big focus will also be on making 'self-improving' AI agents, which can learn and make themselves better without constant human help. Finding scalable approaches to merge originality within feedback collecting will improve Reinforcement Learning from Human Feedback (RLHF). AI models will get better with new designs, simpler numbers (quantification), and by cutting out unneeded parts (pruning). Even simpler numbers, like 1-bit or sub-2-bit, will be used to make them work much faster.

## 13. Conclusion

There have been considerable advances in the field of natural language processing and related domains such as text generation, machine translation, and conversational agents due to the emergence of LLMs. This is a direct result of Instruction Tuning and RLHF, with prompting being more focused on human-directed tasks. Techniques like Few-Shot/Zero-Shot Learning and the use of Multimodal Models allow LLMs to perform intricate cross-modal tasks with little extra training. Unfortunately, building systems that call for Fairness, Transparency, and Trustworthy systems is a challenge that requires further research from the field of Sustainable Designed systems, Sustainable Data Quality, and Bias Mitigation. Future expected efforts are most likely to be aimed at the most pressing issues of the field of energy efficiency, misinformation, and irresponsibly scaling systems. Achieving a balance between complexity, progress and societal impact will be crucial in the field of LLMs.

The future is incredibly promising, and LLMs are likely to continue to push the cutting edge of technology. The thoughtful application of LLMs is likely to result in the automation of complex tasks with a positive impact on the quality of large numbers of lives. The use of LLMs is likely to be positive in the tripling of complex tasks and the automation of technology of human interaction to new levels.

## 14. References

[1] A. Vaswani et al., “Attention Is All You Need,” Aug. 02, 2023, arXiv: arXiv:1706.03762. doi: 10.48550/arXiv.1706.03762.

[2] B. Wang et al., “Towards Understanding Chain-of-Thought Prompting: An Empirical Study of What Matters,” in Proceedings of the 61st Annual Meeting of the Association for Computational Linguistics (Volume 1: Long Papers), Toronto, Canada: Association for Computational Linguistics, 2023, pp. 2717–2739. doi: 10.18653/v1/2023.acl-long.153.

[3] J. Wei et al., “Chain-of-Thought Prompting Elicits Reasoning in Large Language Models,” 2022, arXiv. doi: 10.48550/ARXIV.2201.11903.

[4] G.-G. Lee, E. Latif, X. Wu, N. Liu, and X. Zhai, “Applying large language models and chain-of-thought for automatic scoring,” Comput. Educ. Artif. Intell., vol. 6, p. 100213, Jun. 2024, doi: 10.1016/j.caeai.2024.100213.

[5] Z. Sprague et al., “To CoT or not to CoT? Chain-of-thought helps mainly with math and symbolic reasoning,” 2024, arXiv. doi: 10.48550/ARXIV.2409.12183.

[6] S. Zhang et al., “Instruction Tuning for Large Language Models: A Survey,” 2023, arXiv. doi: 10.48550/ARXIV.2308.10792.

[7] Z. Jiang et al., “Instruction-tuned Language Models are Better Knowledge Learners,” in Proceedings of the 62nd Annual Meeting of the Association for Computational Linguistics (Volume 1: Long Papers), Bangkok, Thailand: Association for Computational Linguistics, 2024, pp. 5421–5434. doi: 10.18653/v1/2024.acl-long.296.

[8] J. Wei et al., “Finetuned Language Models Are Zero-Shot Learners,” 2021, arXiv. doi: 10.48550/ARXIV.2109.01652.

[9] X. Wu et al., “From Language Modeling to Instruction Following: Understanding the Behavior Shift in LLMs after Instruction Tuning,” 2023, arXiv. doi: 10.48550/ARXIV.2310.00492.

[10] N. Mehrabi, F. Morstatter, N. Saxena, K. Lerman, and A. Galstyan, “A Survey on Bias and Fairness in Machine Learning,” ACM Comput. Surv., vol. 54, no. 6, Art. no. 6, Jul. 2022, doi: 10.1145/3457607.

[11] Z. Obermeyer, B. Powers, C. Vogeli, and S. Mullainathan, “Dissecting racial bias in an algorithm used to manage the health of populations,” Science, vol. 366, no. 6464, Art. no. 6464, Oct. 2019, doi: 10.1126/science.aax2342.

[12] E. M. Bender, T. Gebru, A. McMillan-Major, and S. Shmitchell, “On the Dangers of Stochastic Parrots: Can Language Models Be Too Big? □,” in Proceedings of the 2021 ACM Conference on Fairness, Accountability, and Transparency, Virtual Event Canada: ACM, Mar. 2021, pp. 610–623. doi: 10.1145/3442188.3445922.

[13] L. Floridi and J. Cowls, “A Unified Framework of Five Principles for AI in Society,” Harv. Data Sci. Rev.., Jun. 2019, doi: 10.1162/99608f92.8cd550d1.

[14] R. K. E. Bellamy et al., “AI Fairness 360: An Extensible Toolkit for Detecting, Understanding, and Mitigating Unwanted Algorithmic Bias,” 2018, arXiv. doi: 10.48550/ARXIV.1810.01943.

[15] J. Wexler, M. Pushkarna, T. Bolukbasi, M. Wattenberg, F. Viegas, and J. Wilson, “The What-If Tool: Interactive Probing of Machine Learning Models,” IEEE Trans. Vis. Comput. Graph., pp. 1–1, 2019, doi: 10.1109/TVCG.2019.2934619.

[16] H. Weerts, M. Dudík, R. Edgar, A. Jalali, R. Lutz, and M. Madaio, “Fairlearn: Assessing and Improving Fairness of AI Systems,” 2023, arXiv. doi: 10.48550/ARXIV.2303.16626.

[17] P. Saleiro et al., “Aequitas: A Bias and Fairness Audit Toolkit,” 2018, arXiv. doi: 10.48550/ARXIV.1811.05577.

[18] V. Arya et al., “AI Explainability 360: Impact and Design,” 2021, doi: 10.48550/ARXIV.2109.12151.

[19] R. Kumar, H. Kumar, and K. Shalini, “Detecting and Mitigating Bias in LLMs through Knowledge Graph-Augmented Training,” 2025, arXiv. doi: 10.48550/ARXIV.2504.00310.

[20] M. K. Scheuerman, J. M. Paul, and J. R. Brubaker, “How Computers See Gender: An Evaluation of Gender Classification in Commercial Facial Analysis Services,” Proc. ACM Hum.-Comput. Interact., vol.. 3, no. CSCW, Art. no. CSCW, Nov. 2019, doi: 10.1145/3359246.

[21] S. Caton and C. Haas, “Fairness in Machine Learning: A Survey,” ACM Comput. Surv., vol. 56, no. 7, Art. no. 7, Jul. 2024, doi: 10.1145/3616865.

[22] I. D. Raji et al., “Closing the AI accountability gap: defining an end-to-end framework for internal algorithmic auditing,” in Proceedings of the 2020 Conference on Fairness, Accountability, and Transparency, Barcelona, Spain: ACM, Jan. 2020, pp. 33–44. doi: 10.1145/3351095.3372873.

[23] J. A. Mattu, Jeff Larson, Lauren Kirchner, Surya, “Machine Bias,” ProPublica. Accessed: Feb. 21, 2025. [Online]. Available: https://www.propublica.org/article/machine-bias-risk-assessments-in-criminal-sentencing

[24] T. Davidson, D. Warmsley, M. Macy, and I. Weber, “Automated Hate Speech Detection and the Problem of Offensive Language,” Mar. 11, 2017, arXiv: arXiv:1703.04009. doi: 10.48550/arXiv.1703.04009.

[25] R. S. Baker and A. Hawn, “Algorithmic Bias in Education,” Int. J. Artif. Intell. Educ., vol. 32, no. 4, Art. no. 4, Dec. 2022, doi: 10.1007/s40593-021-00285-9.

[26] B. H. Zhang, B. Lemoine, and M. Mitchell, “Mitigating Unwanted Biases with Adversarial Learning,” in Proceedings of the 2018 AAAI/ACM Conference on AI, Ethics, and Society, New Orleans, LA, USA: ACM, Dec. 2018, pp. 335–340. doi: 10.1145/3278721.3278779.

[27] M. Hardt, E. Price, and N. Srebro, "Equality of Opportunity in Supervised Learning," 2016, arXiv. doi: 10.48550/ARXIV.1610.02413.

[28] C. Dwork, M. Hardt, T. Pitassi, O. Reingold, and R. Zemel, "Fairness through awareness," in Proceedings of the 3rd Innovations in Theoretical Computer Science Conference, Cambridge, Massachusetts: ACM, Jan. 2012, pp. 214–226. doi: 10.1145/2090236.2090255.

[29] B. Benbouzid, "Fairness in machine learning from the perspective of sociology of statistics: How machine learning is becoming scientific by turning its back on metrological realism," in 2023 ACM Conference on Fairness, Accountability, and Transparency, Chicago, IL, USA: ACM, Jun. 2023, pp. 35–43. doi: 10.1145/3593013.3593974.

[30] S. Höhn, B. Migge, D. Dippold, B. Schneider, and S. Mauw, "Language Ideology Bias in Conversational Technology," in Chatbot Research and Design, vol. 14524, A. Følstad, T. Araujo, S. Papadopoulos, E. L.-C. Law, E. Luger, M. Goodwin, S. Hobert, and P. B. Brandtzaeg, Eds., in Lecture Notes in Computer Science, vol. 14524. , Cham: Springer Nature Switzerland, 2024, pp. 133–148. doi: 10.1007/978-3-031-54975-5_8.

[31] A. Radford et al., "Learning Transferable Visual Models From Natural Language Supervision," Feb. 26, 2021, arXiv: arXiv:2103.00020. doi: 10.48550/arXiv.2103.00020.

[32] J.-B. Alayrac et al., "Flamingo: a Visual Language Model for Few-Shot Learning," Nov. 15, 2022, arXiv: arXiv:2204.14198. doi: 10.48550/arXiv.2204.14198.

[33] P. Xu, X. Zhu, and D. A. Clifton, "Multimodal Learning with Transformers: A Survey," May 10, 2023, arXiv: arXiv:2206.06488. doi: 10.48550/arXiv.2206.06488.

[34] A. Agostinelli et al., "MusicLM: Generating Music From Text," Jan. 26, 2023, arXiv: arXiv:2301.11325. doi: 10.48550/arXiv.2301.11325.

[35] J. P. Papa and J. M. R. S. Tavares, "Editorial of the special section on CIARP 2021," Pattern Recognit. Lett., vol. 163, p. 182, Nov. 2022, doi: 10.1016/j.patrec.2022.10.004.

[36] J. Devlin, M.-W. Chang, K. Lee, and K. Toutanova, "BERT: Pre-training of Deep Bidirectional Transformers for Language Understanding," May 24, 2019, arXiv: arXiv:1810.04805. doi: 10.48550/arXiv.1810.04805.

[37] A. Khan, L. Asmatullah, A. Malik, S. Khan, and H. Asif, "A Survey on Self-supervised Contrastive Learning for Multimodal Text-Image Analysis," 2025, arXiv. doi: 10.48550/ARXIV.2503.11101.

[38] H. Inaguma et al., "UnitY: Two-pass Direct Speech-to-speech Translation with Discrete Units," May 26, 2023, arXiv: arXiv:2212.08055. doi: 10.48550/arXiv.2212.08055.

[39] T. B. Brown et al., "Language models are few-shot learners," in Proceedings of the 34th International Conference on Neural Information Processing Systems, in NIPS '20. Red Hook, NY, USA: Curran Associates Inc., 2020.

[40] C. Finn, P. Abbeel, and S. Levine, “Model-Agnostic Meta-Learning for Fast Adaptation of Deep Networks,” 2017, arXiv. doi: 10.48550/ARXIV.1703.03400.

[41] L. Reynolds and K. McDonell, “Prompt Programming for Large Language Models: Beyond the Few-Shot Paradigm,” in Extended Abstracts of the 2021 CHI Conference on Human Factors in Computing Systems, in CHI EA’21. New York, NY, USA: Association for Computing Machinery, May 2021, pp. 1–7. doi: 10.1145/3411763.3451760.

[42] A. Hanafi, M. Saad, N. Zahran, R. J. Hanafy, and M. E. Fouda, “A Comprehensive Evaluation of Large Language Models on Mental Illnesses,” 2024, arXiv. doi: 10.48550/ARXIV.2409.15687.

[43] S. Karpurapu et al., “Comprehensive Evaluation and Insights Into the Use of Large Language Models in the Automation of Behavior-Driven Development Acceptance Test Formulation,” IEEE Access, vol. 12, pp. 58715–58721, 2024, doi: 10.1109/ACCESS.2024.3391815.

[44] T. Ahmed and P. Devanbu, “Few-shot training LLMs for project-specific code-summarization,” in Proceedings of the 37th IEEE/ACM International Conference on Automated Software Engineering, Rochester, MI, USA: ACM, Oct. 2022, pp. 1–5. doi: 10.1145/3551349.3559555.

[45] M. Bhattarai, J. Estrada Santos, S. Jones, A. Biswas, B. Alexandrov, and D. O’Malley, “Enhancing Code Translation in Language Models with Few-Shot Learning via Retrieval-Augmented Generation,” in 2024 IEEE High Performance Extreme Computing Conference ; 2024-09-23 - 2024-09-27, US DOE, Sep. 2024. doi: 10.2172/2447962.

[46] DeepSeek-AI et al., “DeepSeek-R1: Incentivizing Reasoning Capability in LLMs via Reinforcement Learning,” 2025, arXiv. doi: 10.48550/ARXIV.2501.12948.

[47] A. S. Nipu, K. M. S. Islam, and P. Madiraju, “How Reliable AI Chatbots are for Disease Prediction from Patient Complaints?” 2024, arXiv. doi: 10.48550/ARXIV.2405.13219.

[48] B. S. Glicksberg et al., “Evaluating the accuracy of a state-of-the-art large language model for prediction of admissions from the emergency room,” J. Am. Med. Inform. Assoc., vol. 31, no. 9, pp. 1921–1928, Sep. 2024, doi: 10.1093/jamia/ocae103.

[49] M. A. Arshad et al., “Leveraging Vision Language Models for Specialized Agricultural Tasks,” 2024, arXiv. doi: 10.48550/ARXIV.2407.19617.

[50] L. Foppiano, G. Lambard, T. Amagasa, and M. Ishii, “Mining experimental data from materials science literature with large language models: an evaluation study,” Sci. Technol. Adv. Mater. Methods, vol. 4, no. 1, p. 2356506, Dec. 2024, doi: 10.1080/27660400.2024.2356506.

[51] G. Chen et al., “Zero-Shot Cross-Lingual Transfer of Neural Machine Translation with Multilingual Pretrained Encoders,” in Proceedings of the 2021 Conference on Empirical Methods in Natural Language Processing, Online and Punta Cana, Dominican Republic: Association for Computational Linguistics, 2021, pp. 15–26. doi: 10.18653/v1/2021.emnlp-main.2.

[52] A. Zhu, X. Lu, X. Bai, S. Uchida, B. K. Iwana, and S. Xiong, “Few-Shot Text Style Transfer via Deep Feature Similarity,” IEEE Trans. Image Process., vol. 29, pp. 6932–6946, 2020, doi: 10.1109/TIP.2020.2995062.

[53] T. Kojima, S. S. Gu, M. Reid, Y. Matsuo, and Y. Iwasawa, “Large language models are zero-shot reasoners,” in Proceedings of the 36th International Conference on Neural Information Processing Systems, in NIPS ’22. Red Hook, NY, USA: Curran Associates Inc., 2022.

[54] A. Madotto, Z. Lin, G. I. Winata, and P. Fung, “Few-Shot Bot: Prompt-Based Learning for Dialogue Systems,” 2021, arXiv. doi: 10.48550/ARXIV.2110.08118.

[55] P. Zhao and S. Jin, “Fewshing: A Few-Shot Learning Approach to Phishing Email Detection,” in 2024 IEEE 4th International Conference on Software Engineering and Artificial Intelligence (SEAI), Xiamen, China: IEEE, Jun. 2024, pp. 371–375. doi: 10.1109/SEAI62072.2024.10674290.

[56] D. Y. Demirel and M. T. Sandikkaya, “Web-Based Anomaly Detection Using Zero-Shot Learning With CNN,” IEEE Access, vol. 11, pp. 91511–91525, 2023, doi: 10.1109/ACCESS.2023.3303845.

[57] R. Sarkar, A. Kr. Ojha, J. Megaro, J. Mariano, V. Herard, and J. P. McCrae, “Few-shot and Zero-shot Approaches to Legal Text Classification: A Case Study in the Financial Sector,” in Proceedings of the Natural Legal Language Processing Workshop 2021, Punta Cana, Dominican Republic: Association for Computational Linguistics, 2021, pp. 102–106. doi: 10.18653/v1/2021.nllp-1.10.

[58] S. Sivarajkumar and Y. Wang, “Evaluation of Healthprompt for Zero-shot Clinical Text Classification,” in 2023 IEEE 11th International Conference on Healthcare Informatics (ICHI), Houston, TX, USA: IEEE, Jun. 2023, pp. 492–494. doi: 10.1109/ICHI57859.2023.00081.

[59] “In what applications is RLHF implementation most effective?” Accessed: Feb. 21, 2025. [Online]. Available: https://www.deepchecks.com/question/best-applications-for-rlhf/

[60] “Reinforcement learning from human feedback,” Wikipedia. Feb. 20, 2025. Accessed: Feb. 21, 2025. [Online]. Available: https://en.wikipedia.org/w/index.php?title=Reinforcement_learning_from_human_feedback&oldid=1276658955

[61] “RLHF Services for Advanced AI Training - Macgence.” Accessed: Feb. 21, 2025. [Online]. Available: https://macgence.com/rlhf

[62] L. Ouyang et al., “Training language models to follow instructions with human feedback,” Mar. 04, 2022, arXiv: arXiv:2203.02155. doi: 10.48550/arXiv.2203.02155.

[63] “What is RLHF? - Reinforcement Learning from Human Feedback Explained - AWS.” Accessed: Feb. 21, 2025. [Online]. Available: https://aws.amazon.com/what-is/reinforcement-learning-from-human-feedback/

[64] "Introducing ChatGPT." Accessed: Feb. 21, 2025. [Online]. Available: https://openai.com/index/chatgpt/

[65] "AI Model Scaling Isn't Over: It's Entering a New Era." Accessed: May 23, 2025. [Online]. Available: https://aibusiness.com/language-models/ai-model-scaling-isn-t-over-it-s-entering-a-new-era

[66] "AI Software Development Best Practices: Scalability & Performance - nCube." Accessed: May 23, 2025. [Online]. Available: https://ncube.com/ai-software-development-best-practices-for-scalability-and-performance

[67] N. Arora, "Exploring the Latest in AI: The Revolutionary LLaMA Model by Meta," Medium. Accessed: May 24, 2025. [Online]. Available: https://medium.com/@nitisharora41/imooexploring-the-latest-in-ai-the-revolutionary-llama-model-by-meta-b654a1a5f2d4

[68] V. Samborska, "Scaling up: how increasing inputs has made artificial intelligence more capable," Our World Data, Jan. 2025, Accessed: May 24, 2025. [Online]. Available: https://ourworldindata.org/scaling-up-ai

[69] S. Rallabandi, "Google's Pathways to Language Model (PaLM): Scaling to New Heights in AI Understanding," Medium. Accessed: May 24, 2025. [Online]. Available: https://medium.com/@sreeku.ralla/googles-pathways-to-language-model-palm-scaling-to-new-heights-in-ai-understanding-c900b0e87c22

[70] J. Kaplan et al., "Scaling Laws for Neural Language Models," 2020, arXiv. doi: 10.48550/ARXIV.2001.08361.

[71] M. Saroufim et al., "NeurIPS 2023 LLM Efficiency Fine-tuning Competition," 2025, arXiv. doi: 10.48550/ARXIV.2503.13507.

[72] A. Khan et al., "A survey of the Vision Transformers and their CNN-Transformer based Variants," 2023, doi: 10.48550/ARXIV.2305.09880.

[73] "Anthropic says its new Opus 4.6 platform found over 500 previously unknown high-severity security flaws in open-source libraries during testing," *TechRadar*, Feb. 2026. [Online]. Available: https://www.techradar.com/pro/security/anthropic-says-its-new-opus-4-6-platform-found-over-500-previously-unknown-high-severity-security-flaws-in-open-source-libraries-during-testing